\documentclass[10pt,twocolumn,twoside]{IEEEtran}

\usepackage{amsmath,amsfonts}
\usepackage{amsthm,amsmath,amssymb}
\usepackage{mathrsfs}
\usepackage{array}
\usepackage{textcomp}
\usepackage{stfloats}
\usepackage{url}
\usepackage{verbatim}
\usepackage{graphicx}
\graphicspath{ {./images/} }
\usepackage{epstopdf}

\usepackage[table]{xcolor}
\def\BibTeX{{\rm B\kern-.05em{\sc i\kern-.025em b}\kern-.08em
    T\kern-.1667em\lower.7ex\hbox{E}\kern-.125emX}}
\usepackage{balance}
\usepackage[labelformat=simple]{subcaption}

\usepackage{multirow}
\usepackage{booktabs}
\usepackage{xcolor}
\usepackage{braket}
\usepackage{hyperref}
\usepackage{pifont}
\usepackage{tikz}
\usepackage{graphicx}
\usepackage{algorithm}
\usepackage{makecell}
\usepackage[noend]{algpseudocode}
\makeatletter
\renewcommand{\maketag@@@}[1]{\hbox{\m@th\normalsize\normalfont#1}}%
\makeatother

\begin{document}

\title{DesCLIP: Robust Continual Learning via General Attribute Descriptions for VLM-Based Visual Recognition}
\author{Chiyuan He, Zihuan Qiu, Fanman Meng, \textit{Member, IEEE}, Linfeng Xu, \textit{Member, IEEE}, \\ Qingbo Wu, \textit{Member, IEEE}, Hongliang Li, \textit{Senior Member, IEEE}

 \thanks{Corresponding author: Fanman Meng. The authors are with the School of Information and Communication Engineering, University of Electronic Science and Technology of China, Chengdu 611731, China (e-mail: \{cyhe;zihuanqiu\}@std.uestc.edu.cn; \{fmmeng;lfxu;qbwu;hlli\}@uestc.edu.cn. This work was supported in part by the National Science and Technology Major Project under Grant 2021ZD0112001, the National Natural Science Foundation of China under Grant (62271119, 62071086), the Key Research and Development Project of Hainan Province under Grant ZDYF2024(LALH)003, and the Natural Science Foundation of Sichuan Province under Grant (2023NSFSC1972, 2025ZNSFSC0475).
} 
}

\maketitle
\begin{abstract}
Continual learning of vision-language models (VLMs) focuses on leveraging cross-modal pretrained knowledge to incrementally adapt to expanding downstream tasks and datasets, while tackling the challenge of knowledge forgetting. Existing research often focuses on connecting visual features with specific class text in downstream tasks, overlooking the latent relationships between general and specialized knowledge. Our findings reveal that forcing models to optimize inappropriate visual-text matches exacerbates forgetting of VLM's recognition ability. To tackle this issue, we propose DesCLIP, which leverages general attribute (GA) descriptions to guide the understanding of specific class objects, enabling VLMs to establish robust \textit{vision-GA-class} trilateral associations rather than relying solely on \textit{vision-class} connections. Specifically, we introduce a language assistant to generate concrete GA description candidates via proper request prompts. Then, an anchor-based embedding filter is designed to obtain highly relevant GA description embeddings, which are leveraged as the paired text embeddings for visual-textual instance matching, thereby tuning the visual encoder. Correspondingly, the class text embeddings are gradually calibrated to align with these shared GA description embeddings. Extensive experiments demonstrate the advancements and efficacy of our proposed method, with comprehensive empirical evaluations highlighting its superior performance  in VLM-based recognition compared to existing continual learning methods.
\end{abstract}


\begin{figure}[t]
  \centering
  \includegraphics[width=0.75\linewidth]{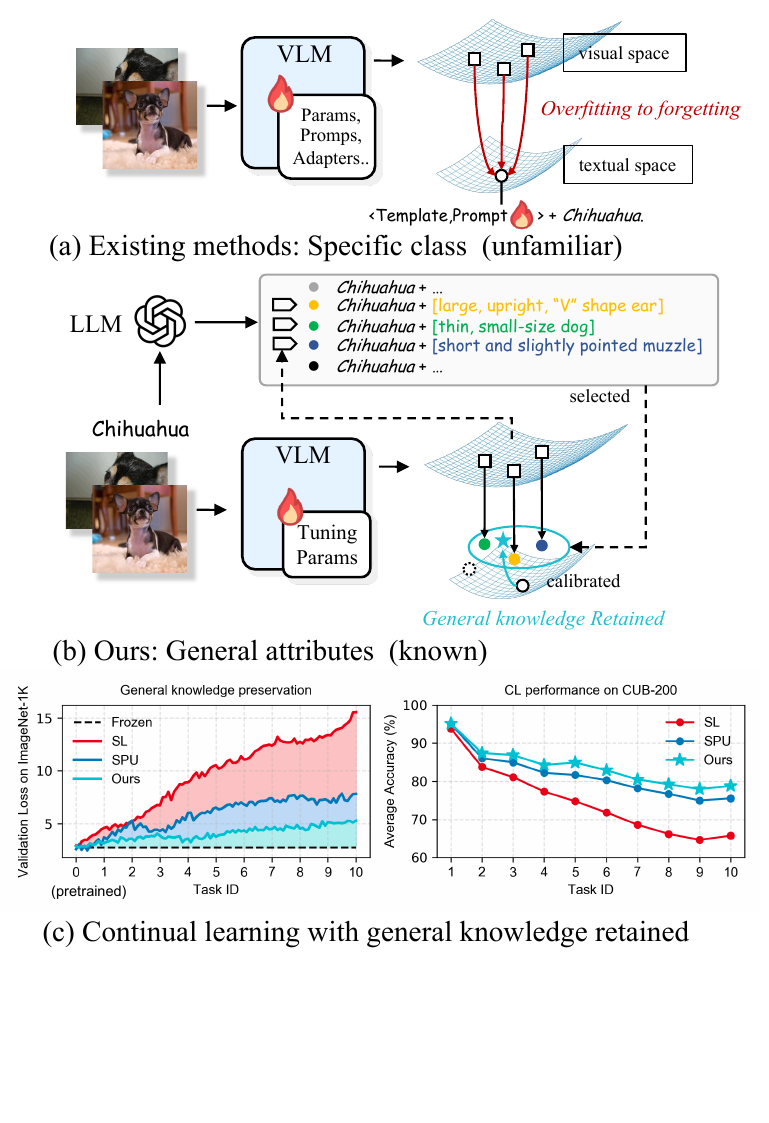}
  \vspace{-4.5em}
  \caption{(a) Existing methods: learning to match unfamiliar specific classes leads to a risk of forgetting. (b) Ours: learning to construct connections with highly relevant general attributes and gradually calibrate class-text embeddings, thereby significantly mitigating the forgetting caused by fitting unfamiliar class knowledge in downstream continual tasks. (c) The maintained low loss on the reference validation set demonstrates that our approach preserves the substantial potential of CL by retaining the pretrained general knowledge, thereby achieving optimal performance, compared to SL \cite{zhang2023slca} and SPU \cite{zhang2024overcoming}.}
  \label{fig:intro}
\end{figure}

\section{Introduction}
\IEEEPARstart{I}{n} recent years, deep models pretrained on large-scale datasets have achieved remarkable success across visual, linguistic, and multi-modal domains. Pretrained vision-language models (VLMs), exemplified by CLIP \cite{radford2021learning} and ALIGN \cite{jia2021scaling}, have demonstrated substantial promise in handling open-vocabulary tasks. Despite their strong zero-shot capabilities in common domains, VLMs often underperform on specialized tasks, such as distinguishing low-quality images or identifying fine-grained object categories. Consequently, significant efforts \cite{zhou2022learning,zhou2022conditional,yu2023task,10171397, yao2023visual,yao2024tcp} have focused on adapting VLMs on downstream datasets to adapt to these new tasks. However, as the demand for and volume of data continue to grow, incorporating previous models and data for joint training imposes substantial storage and computational overhead. Considering the significant cost of repeatedly training foundation models, exploring continual learning (CL) becomes particularly valuable in this context. Nowadays, CL research increasingly shifts from learning from scratch to building upon a generalist “expert” already equipped with broad knowledge, and progressively adapting it to the demands of downstream tasks. This paradigm is especially crucial for enabling continual exploration in open-world scenarios.

Recently, Zheng et al. \cite{zheng2023preventing} and Zhang et al. \cite{zhang2024overcoming} have highlighted the risk of losing existing generalization capabilities when adjusting VLMs with generic knowledge to specialized domain models. This adjustment may result in the model losing effectiveness on prior tasks and limiting their potential for optimization on subsequent tasks. This phenomenon, known as catastrophic forgetting in the field of continual learning, is particularly pronounced on VLMs. Unlike conventional scenarios \cite{li2017learning,10347466,10520827,9899753}, catastrophic forgetting of VLMs impacts not only the task-specific knowledge of previously learned tasks but also the extensive pretrained knowledge, presenting significant challenges in adjusting VLMs for continual tasks.


Over the past few years, research \cite{wang2023attriclip, zhang2024overcoming, xiang2024language, yu2024boosting, zheng2023preventing} has explored how VLMs can adapt to CL recognition tasks. Conventional adapting approaches \cite{zhou2022learning, zhou2022conditional, yao2023visual, pantazis2022svl, xin2024vmt} have shown limited effectiveness in adapting pretrained VLMs for incremental tasks. This limitation is primarily due to their reliance on shared structures and prompts across all tasks, which often results in forgetting of previously learned knowledge when accommodating new information. To address this issue, Wang et al. \cite{wang2023attriclip} proposed AttriCLIP, a sparse prompt selection mechanism that selects abstract attributes with high visual relevance to enrich textual hint prompts. However, these prompts depend solely on image conditions and lack association with class-relevant information, which restricts the effectiveness. Other works \cite{wang2022sparcl, konishi2023parameter, zhang2024overcoming} have focused on selectively updating parameters of pretrained models to better support incremental continual tasks. More recently, approaches incorporating additional structures \cite{xiang2024language, yu2024boosting} with sparse mechanisms have shown promise in mitigating conflicts with prior knowledge and alleviating forgetting of VLMs. Additionally, Zheng et al. \cite{zheng2023preventing} and Yu et al. \cite{yu2025select} use the additional reference datasets to perform knowledge distillation, effectively mitigating the forgetting of generic knowledge. 

\begin{figure}[t]
  \centering
  \includegraphics[width=0.8\linewidth]{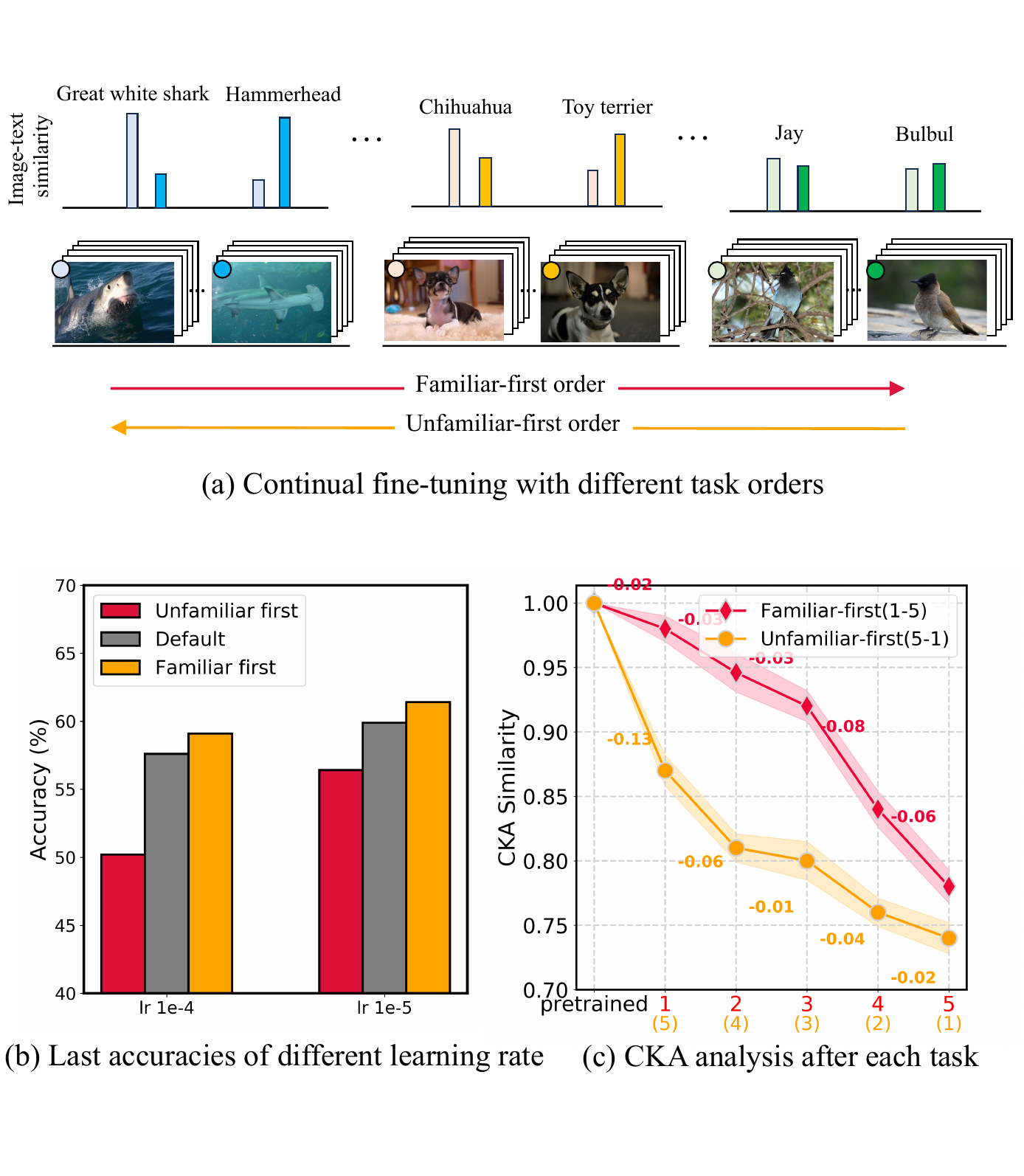}
  \vspace{-2.5em}
  \caption{(a) Continual fine-tuning with different task orders: familiar-first order and unfamiliar-first order. (b) Initially matching unfamiliar class texts leads to more severe forgetting, which negatively impacts the learning of subsequent tasks, resulting in overall poorer CL performance. (c) CKA \cite{kornblith2019similarity} similarities of representation compared to a pretrained CLIP. Task 1-5 represents `familiar to unfamiliar'. Learning to match unfamiliar classes further disrupts the integrity of pretrained representations.}
  \label{fig:intro_exp}
\end{figure}

Although these studies have demonstrated some effectiveness in mitigating knowledge forgetting of VLMs, they can largely be regarded as adaptations of traditional and pretrained-vision-model-based CL methods, tailored to fit the VLM framework. Yet, they fail to fully exploit the robust cross-modal knowledge associations established during the VLM pretraining phase. For instance, these approaches overlook the visual-textual associations established via pretraining, instead relying heavily on rudimentary textual prompts (e.g., \texttt{A photo of a <class~name>, <prompts>+<class~name>}) to correlate with visual information. This introduces a risk of forgetting in unfamiliar downstream tasks (Fig. \ref{fig:intro} (a)). For example, CLIP may struggle with unfamiliar categories like ``European garden spider," resulting in a weak correspondence between its visual and textual representations. Forcing this association can lead to overfitting, which in turn accelerates the forgetting of pretrained and previously learned knowledge. In Fig. \ref{fig:intro_exp}, we demonstrate that beginning with unfamiliar tasks (characterized by low image-text similarity confidence) or from familiar tasks yields markedly different evaluation results on selected subtasks of ImageNet \cite{deng2009imagenet}. Fig. \ref{fig:intro_exp} (b) underscores that forcibly aligning unfamiliar visual-text pairs hinders knowledge retention of VLMs, detracting from subsequent task learning. Additionally, Fig. \ref{fig:intro_exp} (c) uses centered kernel alignment (CKA) \cite{kornblith2019similarity} to analyze representation similarity of continually tuned CLIP to a pretrained CLIP. Notably, learning to match vision-language knowledge unfamiliar to CLIP consistently leads to a substantial decline in CKA at every stage, further disrupting the integrity of pretrained representations and hindering the preservation of general knowledge.

Another line of research~\cite{pratt2023does, menon2022visual, mirza2023lafter, esfandiarpoor2023follow, saha2024improved, mirza2024meta} has attempted to enrich textual descriptions of specific concepts to enhance the performance of VLMs in downstream recognition tasks. However, these methods often lack explicit interaction with visual features and fail to inject fine-grained, attribute-level descriptions into the model's optimization process in a controllable way.

To overcome these limitations, we emphasize the transfer from generalized knowledge to specialized insights. To our knowledge, existing research has not focused on addressing forgetting by establishing robust associations between general attributes and specialized downstream classes. Our approach guides the continual learning process by exploring the context encoding ability of the language branch, forming strong links between general and specialized knowledge. As shown in Fig. \ref{fig:intro}, we advocate for visual representations that align closely with highly-relevant general attribute (GA) embeddings, which are well-known to VLMs, instead of relying on naive class-text embeddings. This prevents the risk of overfitting to unfamiliar classes, as such overfitting can lead to knowledge forgetting. By gradually calibrating text embeddings to align with shared GA embeddings, we form GA-class associations for these incremental downstream tasks. In essence, we redirect the focus from conventional \textbf{vision-class text} connections to establishing robust \textbf{vision-GA-class} trilateral associations, enabling a more effective knowledge transfer that significantly mitigates forgetting of VLMs in downstream recognition tasks. As shown in Fig. \ref{fig:intro}(c), our method continually supplements GA concepts to guide the VLM in adapting to new downstream tasks. This paradigm better preserves the pretrained knowledge, as evidenced by the ability to maintain a low loss on the reference validation set (i.e., ImageNet-1K\cite{deng2009imagenet}), thereby substantially enhancing the potential of CL scenarios. Building on this capability, we represents a novel VLM-based CL paradigm: by introducing task-agnostic general attribute knowledge, the VLM is encouraged to continually adapt to broad tasks without relying on any information from previous tasks.
In summary, our main contributions are as follows:

\begin{itemize}
    \item We revisit the continual learning of VLM-based recognition, focusing on the incremental transfer from generalized to specialized knowledge. By introducing concrete descriptions of general attributes (GAs), we establish more robust vision-GA-class trilateral associations during downstream incremental phases, effectively mitigating forgetting caused by inappropriate visual-text matching.
    \item We propose an anchor-based embedding filter to identify and retain GA description embeddings highly relevant to visual representations. Building on this, we introduce a GA-Guided progressive visual-textual alignment scheme to guide the learning process.
    \item Our method introduces no additional overhead in terms of model structure, data replay, or feature rehearsal storage. Extensive experiments on 6 datasets demonstrate the exceptional performance of our approach. Thorough ablation studies and analyses further corroborate its effectiveness.
\end{itemize}

\section{Related Work}
\label{sec:Related Work}

\subsection{Continual Learning}
\label{sec:Continual Learning}

Continual Learning (CL) investigates how deep models can incrementally learn knowledge. Existing CL research can be categorized into several types based on the strategies they employ. Among these, regularization-based methods \cite{li2017learning, kirkpatrick2017overcoming, douillard2020podnet} introduce regularization terms during model training to penalize forgetting old knowledge. These regularization terms can either focus on protecting model parameters \cite{kirkpatrick2017overcoming} or on output distributions \cite{li2017learning, rebuffi2017icarl} (e.g., knowledge distillation). Dynamic network-based methods \cite{yan2021dynamically, zhou2022model} aim to learn the model by introducing new structures for new tasks while preserving old knowledge, although this incurs substantial overhead as model parameters increase with the number of tasks. Recently, replay-based methods have become increasingly common. Data replay methods \cite{wang2022anti, zhou2022model} assist models in retaining old knowledge by recalling a small number of real samples. Some methods \cite{zhu2021class,petit2023fetril} recall old knowledge by storing sample features and the distributions of these features. However, replay-based methods introduce storage costs and require repetitive computation for old data. Additionally, studies \cite{wu2025demystifying,cheng2025achieving,hu2024task} have shown that focusing on task-specific null-space projection and gradient orthogonality can lead to steady improvements in the stability–plasticity trade-off of the CL model.

In recent years, studies such as \cite{zhou2024expandable, zhou2025revisiting, wang2022learning, wang2022dualprompt} have predominantly focused on integrating additional components for incremental tasks, such as learnable prompts \cite{wang2022learning, wang2022dualprompt, smith2023coda, gao2024consistent} and adapters \cite{zhou2024expandable, zhou2025revisiting}, into pretrained models. This integration necessitates the development of methods for selecting and evaluating the relevance of these components to ensure both their appropriateness and compatibility with the pretrained model. To further mitigate task interference under this paradigm, Lu et al. \cite{lu2024visual} propose learning prompts in orthogonal subspaces, which helps achieve better task decoupling during continual learning. Building on this idea, Lian et al. \cite{liang2024inflora} extend the subspace-based isolation strategy to the Low-Rank Adaptation (LoRA) \cite{hu2022lora} framework, a parameter-efficient fine-tuning method that introduces trainable low-rank matrices alongside frozen pretrained weights. This extension enables more effective reuse of pretrained parameters across tasks, while still preserving modularity and scalability.


\begin{figure*}[htbp]
  \centering
  \includegraphics[width=0.9\linewidth]{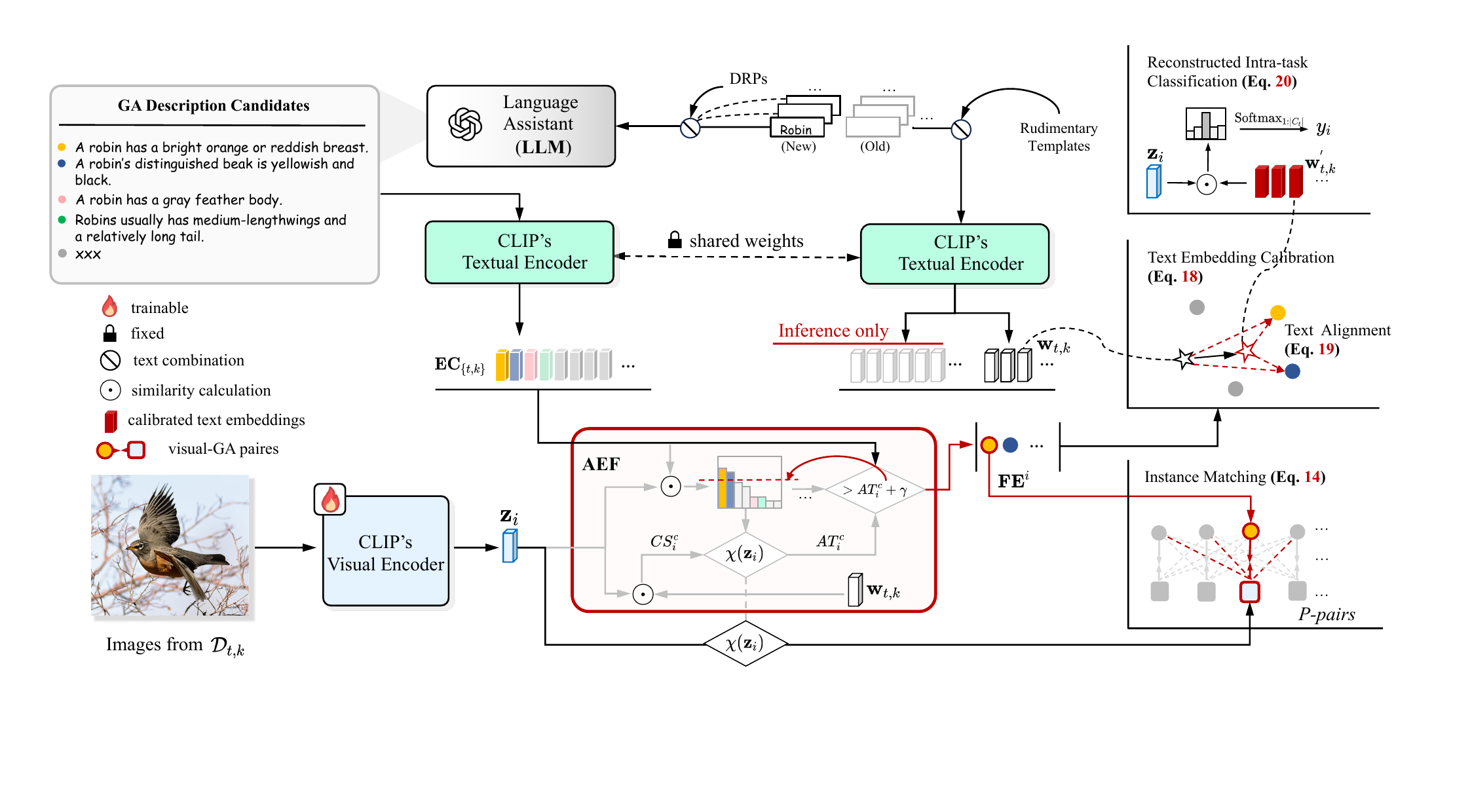}
   \vspace{-4.5em}
  \caption{The overview of our proposed \textbf{DesCLIP}. At each task $t$, language assistant are requested to generate sufficient general attribute description candidates for the classes in the current task, which are then encoded into embeddings via the CLIP's textual encoder. Using the anchor-based embedding filter (\textbf{AEF}), we filter the candidate embeddings by selecting those highly relevant to the visual features of the instances. The filtered embeddings are paired with the instance visual features to compute a class-agnostic instance matching loss. Correspondingly, class text embeddings are calibrated through shift weights to align with these shared filtered embeddings.}
  \label{fig:pipeline}
\end{figure*}

\subsection{Vision-Language Models}

With advancements in pre-training techniques, large-scale foundation models \cite{radford2021learning,li2022blip,kirillov2023segment,openai2023gpt4,alayrac2022flamingo} have significantly impacted the industry. For instance, Vision-Language Models such as Contrastive Language-Image Pretraining (CLIP) \cite{radford2021learning} and Adversarially Learned Inference for Image-Text Matching (ALIGN) \cite{jia2021scaling} have demonstrated remarkable zero-shot capabilities for general tasks. However, despite being pre-trained on over 400 million image-text pairs, CLIP still faces challenges in specific downstream tasks, such as accurately identifying certain types of vehicles and lizards. 

To better adapt VLMs for downstream tasks, especially for visual recognition, various text prompt-based fine-tuning methods \cite{zhou2022learning, zhou2022conditional, 10171397, 10814093} have been proposed, which can enhance VLM performance on specific tasks. In more complex scenarios, learnable prompts can be inserted into intermediate layers \cite{yao2024tcp} to incorporate more sophisticated general knowledge. Moreover, text-semantic-guided learnable prompts \cite{cheng2024disentangled} have also been shown effective in addressing domain generalization problems. Additionally, the integration of adapter structures \cite{li2024graphadapter,pantazis2022svl,xin2024vmt} has also been shown to be an effective strategy. Other approaches \cite{zhang2024concept, wu2024transferring, luo2024cheap, gao2022pyramidclip} focus on the representation alignment of VLMs and aim to improve the transfer of general knowledge. Although these methods demonstrate excellent performance in CLIP transfer tasks, they are inherently unsuitable for incremental learning, as the additional learnable structures cannot effectively mitigate catastrophic forgetting. 

Recently, leveraging the fine-grained knowledge of large language models (LLMs) to enhance the vision-language alignment in VLMs has emerged as a significant research direction. This line of work \cite{pratt2023does, menon2022visual, mirza2023lafter, esfandiarpoor2023follow, saha2024improved, mirza2024meta} is motivated by the observation that VLMs possess not only high-level conceptual understanding but also the capacity to interpret fine-grained visual components. Pratt et al.~\cite{pratt2023does} demonstrated this capability and proposed replacing handcrafted templates with LLM-generated prompts enriched with detailed semantics. DCLIP~\cite{menon2022visual} introduced multiple interpretable descriptors to provide diverse perspectives for class comprehension. Furthermore, Esfandiarpoor et al.~\cite{esfandiarpoor2023follow} proposed FuDD, a two-stage text generation approach that uses discriminative fine-grained descriptions to reduce confusion between visually similar classes. More recently, Mirza et al.~\cite{mirza2024meta} developed MPVR, which employs meta-prompting strategies to produce diverse semantic texts that better support VLMs' understanding of downstream concepts. However, while these methods focus on generating semantically rich textual inputs by leveraging the external knowledge of LLMs, they overlook the importance of controllable model optimization and pay limited attention to preserving transferable knowledge within the model itself.

\subsection{Continual Learning for VLMs}

Investigating the continual learning and adaptation of VLMs for diverse downstream tasks holds significant value, as it reduces data storage requirements and computational redundancy while addressing the challenge of inaccessible previous data. It is crucial to protect the model's generic pretrained knowledge and previously learned knowledge. The full fine-tuning strategies discussed in \ref{sec:Continual Learning} will lead to significant forgetting of pre-trained knowledge, which is a notable distinction between pre-trained foundation models (e.g., CLIP) and small-scale deep models. Additionally, frameworks such as CoOp \cite{zhou2022learning} and CoOpOp \cite{zhou2022conditional} have been shown to have limited adjustment capabilities for VLMs in incremental tasks due to their reliance on shared structures and contextual prompts across all tasks, leading to forgetting old knowledge during the process of fitting new knowledge. To solve this, Wang et al. \cite{wang2023attriclip} introduced AttriCLIP, which establishes a shared attribute bank for all tasks and selects suitable contexts based on visual images to bridge the gap between images and text. To alleviate interference with previously acquired task knowledge in VLMs, Zhou et al. \cite{zhou2025learning} proposed a novel paradigm that combines dynamic projection and fusion mechanisms, facilitating more effective utilization of cross-modal information. Furthermore, Yu et al. \cite{yu2024boosting} introduced a mixture-of-experts Adapter (MoE-Adapter) framework to adaptively manage task-specific knowledge, thereby decoupling the model’s zero-shot capabilities from different specialized task adaptations. From the perspective of parameter sparse updating, efforts from SPG \cite{konishi2023parameter}, SparseCL \cite{wang2022sparcl}, and SPU \cite{zhang2024overcoming} have aimed to update VLM parameters selectively by employing appropriate “important parameter” selection patterns; for example, SPU selects more important parameters for updates based on the gradients accumulated by batches. Additionally, Zheng et al. \cite{zheng2023preventing} and Yu et al. \cite{yu2025select} proposed the use of additional reference datasets to facilitate knowledge distillation in a VLM, effectively mitigating the forgetting of generic knowledge.
\section{Methodology}
\label{sec:Methodology}

\begin{figure*}[t]
  \centering
\includegraphics[width=\linewidth]{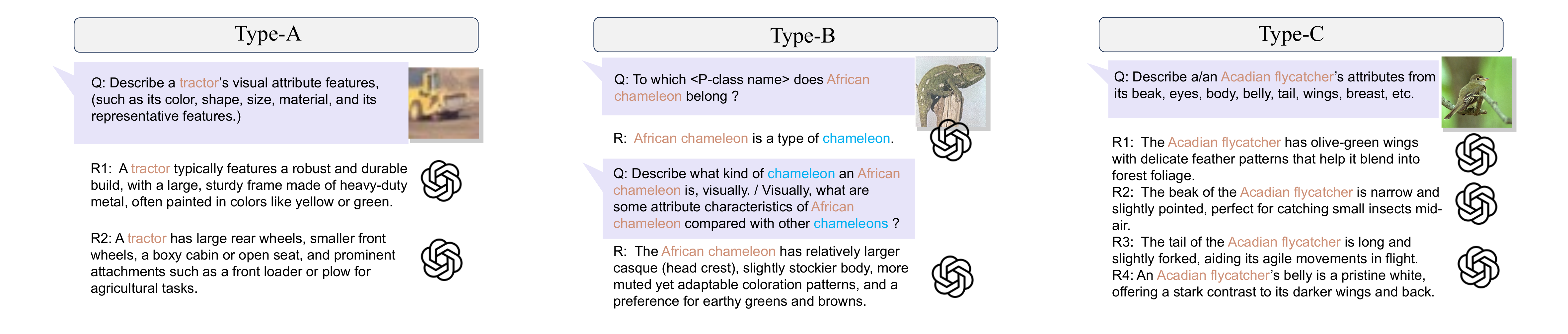}
  \caption{DRP-guided GA description generation with a language assistant.}
  \label{fig:conversation}
\end{figure*}

\subsection{Preliminaries}
\textit{1) Continual Learning Formulation:} A sequence of task datasets is denoted as \( \{\mathcal{D}_1, \mathcal{D}_2, \ldots, \mathcal{D}_T\} \). During training on task \( \mathcal{D}_t \) (\( t \in \{1, 2, \ldots, T\} \)), access to data from previous tasks \( \{\mathcal{D}_1, \mathcal{D}_2, \ldots, \mathcal{D}_{t-1}\} \) is either highly restricted or entirely unavailable. In class-incremental learning (CIL), datasets for different tasks are introduced sequentially. Each task \( t \) is associated with a unique set of classes \( {C}_t = \{{C}_{t,1}, {C}_{t,2}, \ldots, {C}_{t,|{C}_t|}\} \), where \( |{C}_t| \) denotes the number of classes in task \( t \). The classes associated with different tasks are disjoint:
\begin{equation}
    {C}_t \cap {C}_{t'} = \emptyset, \quad \forall t \neq t', \, t, t' \in \{1, 2, \ldots, T\}.
\end{equation}

\textit{2) CLIP for Incremental Tasks:} CLIP \cite{radford2021learning} comprises an image encoder \( \mathcal{F}_{\Theta}(\cdot) \) and a text encoder \( \mathcal{T}(\cdot) \). Specifically, an image \( x \in \mathbb{R}^{H \times W \times 3} \) and a text prompt, referred to as the rudimentary prompt \( \mathbf{RP}_y\), are input into \( \mathcal{F}_{\Theta}(\cdot) \) and \( \mathcal{T}(\cdot) \), respectively, producing a visual embedding \( \mathbf{z} \in \mathbb{R}^D \) and a text embedding \( \mathbf{w}_y \in \mathbb{R}^D \):
\begin{equation}
    \mathbf{z} = \mathcal{F}_{\Theta}(x), \quad \mathbf{w}_y = \mathcal{T}(\mathbf{RP}_y).
\end{equation}
Here, \( \mathbf{RP}_y \) is derived from hand-crafted prompts, typically following a template such as ``\texttt{A photo of a <class~name>}.'' The probability of classifying a test image \( x \) as class \( y_i \) is computed using the softmax function:
\begin{equation}
   p\left(y_{i} \mid x\right) = \frac{\exp\left(\left \langle \frac{\mathbf{z}}{\|\mathbf{z}\|},\frac{\mathbf{w}_{y_i}}{\|\mathbf{w}_{y_i}\|}\right \rangle / \tau\right)}{\sum_{k=1}^{K} \exp\left(\left \langle  \frac{\mathbf{z}}{\|\mathbf{z}\|}, \frac{\mathbf{w}_{y_k}}{\|\mathbf{w}_{y_k}\|} \right \rangle / \tau\right)},
\end{equation}
where \( \tau \) is the temperature parameter, \( \mathbf{w}_{y_k} \) is the class text embedding derived from the rudimentary prompt of the \( k \)-th class, and \( K \) denotes the total number of downstream classes. 

Building on this architecture, ContinualCLIP \cite{thengane2022clip} tackles the challenge of continual learning with a training-free approach. For each new task \( t \), the text embedding set \( \mathbf{W}_t \) is expanded to incorporate embeddings for the new task's classes. At task \( t \), the updated text embedding set is defined as:
\begin{equation}
    \mathbf{W}_t = \bigcup_{j=1}^{t} \bigcup_{k=1}^{|C_{j}|} \mathbf{w}_{j,k},
\end{equation}
where \( \mathbf{w}_{j,k} \) denotes the text embedding for the \( k \)-th class of task \( j \) encountered so far. Consequently, the prediction for a test image \( x \) after task \( t \) is computed as:
\begin{equation}
    p\left(y_{i} \mid x\right) = \frac{\exp\left(\left \langle \frac{\mathbf{z}}{\|\mathbf{z}\|},\frac{\mathbf{w}_{y_i}}{\|\mathbf{w}_{y_i}\|}\right \rangle  / \tau\right)}{\sum_{\mathbf{w}_{j,k} \sim \mathbf{W}_t } \exp\left(\left \langle \frac{\mathbf{z}}{\|\mathbf{z}\|},\frac{\mathbf{w}_{j,k}}{\|\mathbf{w}_{j,k}\|} \right \rangle/ \tau\right)}.
\end{equation}

\subsection{Overview of proposed DesCLIP}
The overall architecture of our proposed framework is shown in Fig.~\ref{fig:pipeline}. Within our framework, the CLIP's textual encoder \( \mathcal{T}(\cdot) \) remains fixed and comprises two input branches: one processes rudimentary prompts derived from basic class names, while the other generates a diverse pool of general attribute (GA) description candidates via a language assistant. To obtain highly relevant visual-GA text pairs, we introduce the anchor-based embedding filter (\textbf{AEF}). The \textbf{AEF} identifies the most relevant attribute description embeddings from the candidate pool with respect to the current visual features. These filtered embeddings are then paired with visual features to compute a class-agnostic instance-matching loss, which is utilized to fine-tune the visual encoder \( \mathcal{F}_{\Theta}(\cdot) \). Concurrently, the text embeddings are gradually calibrated to align with shared attribute embeddings, further enhancing the consistency among representations of vision, GA, and downstream classes.


\subsection{General Attribute Description Generation}

CLIP establishes robust visual-textual associations during the pre-training phase through instance-level contrastive learning. However, most existing research overlooks this foundational capability, conventionally relying on fixed, hand-crafted templates combined with class names as prompts to derive ``prior" classification weights via the textual encoder. Although Wang et al.~\cite{wang2023attriclip} introduced an ``attribute bank" to enable attribute sharing across different task instances, this approach lacks intrinsic relevance to specific classes. For instance, attributes such as ``white" and ``grass" fail to provide meaningful distinctions between classes like ``cat" and ``dog".

To address this limitation, we propose the use of a language assistant to generate rich, contextually relevant attribute descriptions for specific classes. The language assistant utilizes an advanced large language model (LLM) with a generalized understanding of downstream task entities. Drawing inspiration from~\cite{pratt2023does, yi2024leveraging, saha2024improved}, we design a variety of describe-request prompts (DRPs) to guide the language assistant in generating visually relevant attribute descriptions. Examples of basic DRPs include:
    \begin{itemize}
        \item \texttt{Q: Describe what does a/an <class~name> look like?}  
        \item \texttt{Q: Describe a/an <class~name>'s attribute features.}  
        \item \texttt{Q: Describe a/an <class~name>'s outlook features.}  
    \end{itemize}
Additionally, more complex prompts are designed to produce discriminative attribute descriptions, such as:
\begin{itemize}
    \item \texttt{Q: Describe what are some attribute characteristics of <class~name> compared with other <P-class~name>, visually.}  
    \item \texttt{Q: Describe what kind of <P-class~name> a/an <class~name> is, visually.}  
\end{itemize}
Here, \texttt{<P-class~name>} refers to the parent class of {$<class~name>$}. Fine-grained DRPs are also employed for tasks with a known general scope, such as identifying objects within the ``birds" supercategory:
\begin{itemize}
    \item \texttt{Describe a/an <class~name>’s attributes from its beak, eyes, body, belly, tail, wings,
breast, etc.}
\end{itemize}
The DRP-guided general attribute description generation is illustrated in Fig.~\ref{fig:conversation}. The language assistant generates \( n_{dsc} \) attribute description candidates (DCs) for the \( k \)-th class of incremental task \( t \), denoted as:  
\begin{equation}
\mathbf{DC}_{\{t,k\}} = \bigg\{\mathbf{DC}_{\{t,k\},1}, \mathbf{DC}_{\{t,k\},2}, \ldots, \mathbf{DC}_{\{t,k\},n_{dsc}}\bigg\},
\end{equation}
where \( k \in \{1, 2, \ldots, |C_t|\} \). These DCs are then embedded using the textual encoder \( \mathcal{T}(\cdot) \) to produce attribute embedding candidates (ECs):  
\begin{equation}
\begin{split}
\mathbf{EC}_{\{t,k\}} &= \mathcal{T}\Big(\mathbf{DC}_{\{t,k\}}\Big) \\
&= \bigg\{\mathbf{EC}_{\{t,k\},1}, \mathbf{EC}_{\{t,k\},2}, \ldots, \mathbf{EC}_{\{t,k\},n_{dsc}}\bigg\}.
\end{split}
\end{equation}
Each element of \( \mathbf{EC}_{\{t,k\}} \) has the same dimension as the rudimentary text embedding \( \mathbf{w}_{t,k} \).  

\subsection{Anchor-based Embedding Filter}
\label{sec:Anchor-based Embedding Filter}
The generated embedding candidates of GAs do not always align with visual representations due to potential domain discrepancies or unrelated information in the text descriptions produced by the language assistant. To establish robust vision-GA associations, we propose an anchor-based embedding filter (\textbf{AEF}) mechanism to refine the embedding candidates. This mechanism identifies candidates that sufficiently match the visual representations, enabling the construction of approximate image-text pairs tailored to the specific requirements of the task.

For a training sample \((x_i, y_i)\), where \(y_i = c = \{t, k\}\) is assumed, the label \(c\) is considered to correspond to the \(k\)-th class of incremental task \(t\), with \(k \in \{1, 2, \dots, |C_t|\}\). As illustrated in Fig.~\ref{fig:pipeline}, the inputs to \textbf{AEF} include the visual features \(\mathbf{z}_i = \mathcal{F}_{\Theta}(x_i)\), the rudimentary text embedding \(\mathbf{w}_{c} = \mathcal{T}(\textbf{RP}_{y_i})\), and the embedding candidates \(\mathbf{EC}_{c}\). The cosine similarity between the visual features \(\mathbf{z}_i\) and the rudimentary text embedding \(\mathbf{w}_{c}\) is calculated as:  
\begin{equation}
CS_i^c = \left \langle \frac{\mathbf{z}_i}{\|\mathbf{z}_i\|}, \frac{\mathbf{w}_{c}}{\|\mathbf{w}_{c}\|} \right \rangle.
\end{equation}
Subsequently, the similarity scores between the visual features and each embedding candidate in \(\mathbf{EC}_{c}\) are calculated:  
\begin{equation}
\textbf{EC\_S}_{c}^{i,j} = \left \langle \frac{\mathbf{z}_i}{\|\mathbf{z}_i\|}, \frac{\mathbf{EC}_{c,j}}{\|\mathbf{EC}_{c,j}\|} \right \rangle,
\end{equation}
where \(j \in \{1, 2, \dots, n_{dsc}\}\). To mitigate the risk of overfitting in the CLIP’s visual encoder, visual features with low relevance to either class text or GA descriptions should be filtered out, as they are unlikely to support meaningful visual-GA instance matching. Retaining these irrelevant features would introduce noise into the optimization directions during training, thereby increasing the risk of overfitting and accelerating knowledge forgetting. Hence, a condition \(\chi(\mathbf{z}_i)\) is defined to filter visual features as:
\begin{equation}
    \chi(\mathbf{z}_i) = 
\begin{cases} 
1, & \text{if} ~~\max\big(CS_i^c, \max_{j} \textbf{EC\_S}_{c}^{i,j}\big) > \delta_{d}, \\
0, & \text{otherwise}.
\end{cases}
\label{eq_delta_d}
\end{equation}
Here, \(\delta_d\) is a predefined threshold. The similarity between retained visual features and the rudimentary text embedding is used to define the anchor threshold \(AT_i^c\):  
\begin{equation}
AT_i^c = \big\{ CS_i^c \; | \; \chi(\mathbf{z}_i) = 1 \big\}.
\end{equation}
We posit that embedding candidates in \(\mathbf{EC}_c\) exhibiting a similarity score surpassing a threshold \(\gamma\) ($\gamma>0$, relative to the anchor threshold \(AT_i^c\)) are more consistent with the visual features of the current sample. These candidates are filtered as follows:  
\begin{equation}
\label{eq:AEF}
\mathbf{FE}^i = \bigg\{ \textbf{EC}_{c,j} \; | \; \textbf{EC\_S}_{c}^{i,j} > AT_i^c + \gamma \bigg\},
\end{equation}
where $j \in \{1, 2, \dots, n_{dsc}\}$, and sorted with descending order according to $\textbf{EC\_S}_{c}^{i,j}$. To reduce the potential influence of domain discrepancies arising from contextual descriptions, we further restrict the selection process by exclusively retaining attribute description sentences in \(\mathbf{DC}_c\) that explicitly include the class name as a noun (i.e., \texttt{<class~name>+<GA>]}).

\subsection{General Attribute-Guided Progressive Visual-Textual Alignment}

This section introduces the methodology for optimizing the visual and textual branches of CLIP in incremental tasks, leveraging the filtered embeddings identified for relevant training samples. Within the CLIP architecture, the optimization focuses on the visual encoder \(\mathcal{F}_{\Theta}(\cdot)\) (specifically, the initial MLP layers within each Transformer block \cite{zhang2024overcoming}) and the rudimentary text embeddings introduced for the current task.

\textit{1) Instance Matching:} To align the visual features with highly relevant embedded GA descriptions, we select the closest textual representation $\mathbf{h}_i$ as the paired text embedding:
\begin{equation}
\mathbf{h}_i = \mathbf{FE}^i[0].
\end{equation}
The instance matching loss $\mathcal{L}_{\text{IM}}$ is computed across the batch as:
\begin{equation}
    \mathcal{L}_{\text{IM}} = \mathbb{E}_{i \sim P} \Big[ - \log \frac{\mathbf{M}_{i,i}}{\mathbf{M}_{i,i} + \sum_{j \in P, j \neq i} \mathbf{M}_{i,j}} \Big],
\end{equation}
where
\begin{equation}
    \mathbf{M}_{i,j} = \exp\Big(\left \langle \frac{\mathbf{z}_i}{\|\mathbf{z}_i\|}, \frac{\mathbf{h}_j}{\|\mathbf{h}_j\|} \right \rangle / \tilde{\tau}\Big),
\end{equation}
\begin{equation}
    \mathbf{M}_{i,i} = \exp\Big(\left \langle \frac{\mathbf{z}_i}{\|\mathbf{z}_i\|}, \frac{\mathbf{h}_i}{\|\mathbf{h}_i\|} \right \rangle / \tilde{\tau}\Big).
\end{equation}
Here, $\tilde{\tau}$ denotes an elevated temperature scaling factor, defined as $\tilde{\tau} = 10\tau$. The set $P$ represents the indices of valid samples within a batch of size $B$ and is specified as:
\begin{equation}
    P = \Big\{ i \mid i \in \{1, \dots, B\}, \, \chi(\mathbf{z}_i) = 1 \, ,\mathbf{FE}^i \neq \emptyset \Big\},
\end{equation}
where $\chi(\mathbf{z}_i)$ is the condition defined in \ref{sec:Anchor-based Embedding Filter}. For each incremental task \( t \), we tune the model using a contrastive learning framework similar to the pretraining strategy of CLIP \cite{radford2021learning}. To mitigate forgetting, we adhere to a “nearest matching” principle, aligning visual features with GA description embeddings exhibiting higher correlations. This approach minimizes the risk of overfitting visual features to specific class text embeddings, maintaining both generalization and previous knowledge.

\textit{2) Text Embedding Calibration:} General attribute descriptions play a pivotal role in guiding the calibration of text embeddings to achieve better alignment with visual representations. This alignment is particularly crucial because the original rudimentary text embeddings often exhibit weak correlations with visual features in ``unfamiliar'' downstream tasks. Such misalignment can lead to overfitting in the visual branch of the VLM to class labels, thereby exacerbating forgetting. To mitigate this issue, we propose a weight-shifting mechanism that calibrates the rudimentary text embeddings for the classes in task \( t \). This mechanism repositions the text embeddings toward representative attributes shared across the corresponding visual features, fostering stronger alignment between shared general attributes and class-specific text embeddings. Specifically, we define a shifting weight $\mathbf{s}_{t,k}\in \mathbb{R}^D$, and a shift transformation $\Psi(\cdot,\mathbf{s}_{t,k})$ for the calibration of rudimentary text embedding $\mathbf{w}_{t,k}$, where $\{t,k\}$ representing the $k-$th class of incremental task $t$, $k\in \{1,2,...,|C_t|\}$. The calibrated text embedding $\mathbf{w'}_{t,k}$ can be obtained as:
\begin{equation}
\mathbf{w'}_{t,k} = 
\Psi(\mathbf{w}_{t,k},\mathbf{s}_{t,k}) = \frac{\mathbf{w}_{t,k}}{\|\mathbf{w}_{t,k}\|} + \alpha \cdot \mathbf{s}_{t,k}.
\end{equation}
The key to text embedding calibration is to ensure a strong correlation with the visual representations of the class while preventing an excessive focus on any single attribute text. Therefore, $\mathbf{w}'_{t,k}$ should be aligned with $\mathbf{FE}^i$, which is filtered based on the visual features $\mathbf{z}_i$ of the class $c = \{t, k\}$:

    \begin{equation}
    \mathcal{L}_{\text{TA}}^i = \mathbb{E}_{\mathbf{u} \sim \mathbf{FE}^i} 
     \left\| \frac{\mathbf{w}'_{t,k}}{\|\mathbf{w}'_{t,k}\|} - \frac{\mathbf{u}}{\|\mathbf{u}\|} \right\|_2^2.
\end{equation}
Thus, text alignment loss $\mathcal{L}_{\text{TA}}$ of the current batch can be obtained by: $\mathcal{L}_{\text{TA}} = \sum_{i \sim P} \mathcal{L}_{\text{TA}}^i.$ Since, in the context of continual learning, data from previous tasks cannot be revisited during the current task, the weights \(\mathbf{w'}_{t}=\{\mathbf{w'}_{t,1},\mathbf{w'}_{t,2},...,\mathbf{w'}_{t,|C_t|}\}\) are calibrated solely within the scope of the current task \( t \). Consequently, only the shifting weights \(\mathbf{s}_{t}= \{\mathbf{s}_{t,1},\mathbf{s}_{t,2},...,\mathbf{s}_{t,|C_t|}\}\) associated with the current task are learnable, whereas the shifting weights \(\mathbf{s}_{0 : t-1}=\{\mathbf{s}_0,\mathbf{s}_1,...,\mathbf{s}_{t-1}\}\) from previous tasks remain fixed.

\textit{3) Reconstructed Intra-task Classification:} To ensure alignment between text embeddings and the visual branch during optimization, we reconstruct the classification loss for task \( t \). This loss constrains the calibration of text embeddings to remain within the low-loss region of the classification space for the current task, which is a critical prerequisite. Specifically, for each visual feature \(\mathbf{z}_i\) in a batch, its similarity to all calibrated text embeddings for the current task is computed to generate predicted logits. These logits are aligned with the ground truth label \( y_i \), and the prediction classification loss is calculated as:
\begin{equation}
\mathcal{L}_{\text{RIC}}=\mathbb{E}_{i\sim\{1,2,...,B\}}\Bigg[ - \log\frac{\exp\Big(\left \langle \frac{\mathbf{z}_i}{\|\mathbf{z}_i\|}, \frac{\mathbf{w'}_{y_i}}{\|\mathbf{w'}_{y_i}\|}\right \rangle /\tau \Big)}{\sum_{k=1}^{|C_t|}\exp \Big(\left \langle \frac{\mathbf{z}_i}{\|\mathbf{z}_i\|}, \frac{\mathbf{w'}_{t,k}}{\|\mathbf{w'}_{t,k}\|}\right \rangle / \tau \Big)} \Bigg],
\end{equation}
where \(\mathbf{w'}_{y_i}\) represents the calibrated text embedding for the class corresponding to \(\mathbf{z}_i\). 

\textit{4) Optimization:}
To achieve optimal performance, the total loss for batch optimization is defined as:
\begin{equation}
\min_{\Theta, \mathbf{s}_{t}} \mathcal{L} = \lambda_{\text{IM}} \cdot \mathcal{L}_{\text{IM}} + \lambda_{\text{TA}} \cdot \mathcal{L}_{\text{TA}} + \lambda_{\text{RIC}} \cdot \mathcal{L}_{\text{RIC}},
\end{equation}
where \(\lambda_{\text{IM}}\), \(\lambda_{\text{TA}}\), and \(\lambda_{\text{RIC}}\) are balancing factors that control the contributions of the respective loss terms.

\subsection{Inference Stage}
The GA descriptions and embeddings we introduce do not participate in the inference stage of the CLIP after each training phase, which effectively avoids additional storage overhead and eliminates any increase in inference time. Consequently, the model obtained through our method maintains identical parameter size and inference time during the inference stage as the original zero-shot CLIP. In inference stage after task $t$, we leverage the calibrated text embeddings of all seen classes:
    $\mathbf{W'}_t = \bigcup_{j=1}^{t} \bigcup_{k=1}^{|C_{j}|} \mathbf{w'}_{j,k}.$
Hence, the probability of predicting the test image \( x \) as the class \( y_i \) can be expressed as:
\begin{equation}
p\left(y_{i} \mid x\right) = \frac{\exp\left(\left \langle \frac{\mathbf{z}}{\|\mathbf{z}\|} , \frac{\mathbf{w'}_{y_{i}}}{\|\mathbf{w'}_{y_{i}}\|}  \right \rangle / \tau\right)}{\sum_{\mathbf{w}'_{j,k} \sim \mathbf{W'}_t } \exp\left(\left \langle  \frac{\mathbf{z}}{\|\mathbf{z}\|} , \frac{\mathbf{w}'_{j,k}}{\|\mathbf{w}'_{j,k}\|} \right \rangle / \tau\right)}.
\end{equation}

\section{Experiments}
\subsection{Setup}

\textit{1) Datasets: } The evaluation experiments for continual learning are conducted on CIFAR100 \cite{krizhevsky2009learning}, ImageNet-Sub \cite{deng2009imagenet}, CUB-200 \cite{wah2011caltech}, and ImageNet-R \cite{hendrycks2021imagenet}. To further evaluate generalization under `unseen' conditions, we additionally incorporate two out-of-distribution (OOD) datasets: FGVC-Aircraft \cite{maji2013fine} and SKIN-40 \cite{xu2022expert}.

\textit{2) Metrics:} To evaluate the continual learning performance of classification models, we employ two primary metrics: `Last' and `Avg'. `Last' denotes the average accuracy across all classes after the model has completed training on the final task. `Avg' represents the mean incremental accuracy calculated over all tasks the model has learned thus far. In addition, the control set accuracy `C.' \cite{zhang2024overcoming} is employed to evaluate the retention of CLIP's zero-shot generalization knowledge after continual learning, assessed on ImageNet-1K \cite{deng2009imagenet} (1000 classes in total).
\begin{table*}[h]
    \centering
    \caption{
Comparison of different methods on the CIFAR100, ImageNet-Sub, and CUB-200 datasets under various settings with CLIP ViT-L/14. `\textbf{UB}' denotes the upper bound achieved through joint training.}
    \label{tab:comparison}
    \renewcommand{\arraystretch}{1.1}
    \setlength{\tabcolsep}{6pt}
    \resizebox{\textwidth}{!}{%
    
    \begin{tabular}{l|c|cccccc|cccccc|cccccc}
    \toprule[1.2pt]
    \multirow{3}{*}{\textbf{Method}} & \multirow{3}{*}{\textbf{Mem. of Old Tasks}}
    & \multicolumn{6}{c|}{CIFAR100 }
    & \multicolumn{6}{c|}{ImageNet-Sub }
    & \multicolumn{6}{c}{CUB-200 } \\
    & & \multicolumn{2}{c}{$T=5$} & \multicolumn{2}{c}{$T=10$} & \multicolumn{2}{c|}{$T=20$}
      & \multicolumn{2}{c}{$T=5$} & \multicolumn{2}{c}{$T=10$} & \multicolumn{2}{c|}{$T=20$}
      & \multicolumn{2}{c}{$T=5$} & \multicolumn{2}{c}{$T=10$} & \multicolumn{2}{c}{$T=20$} \\
    & & Last & Avg & Last & Avg & Last & Avg
      & Last & Avg & Last & Avg & Last & Avg
      & Last & Avg & Last & Avg & Last & Avg \\
        \midrule
               Seq FT        & \textcolor{purple}{\ding{55}}         & --     & --    & 50.6 & 66.9 & --    & --    & --   & --    & 58.8 & 73.8 & --   & --    & --   & --    & 40.8 & 64.6 & --   & --   \\ 
        LwF \cite{li2017learning}        & \textcolor{purple}{\ding{55}}   & --      & --     & 56.1 & 74.8 & --    & --    & --   & --    & 56.4 & 75.0 & --   & --    & --   & --    & 52.1 & 71.5 & --   & --   \\ 
        EWC \cite{kirkpatrick2017overcoming}        & \textcolor{green!60!black}{\checkmark}Fisher Approx.   & --     & --    & 64.4 & 78.4 & --    & --    & --   & --    & 70.3 & 80.2 & --   & --    & --   & --    & 64.4 & 78.6 & --   & --   \\ 
        AdaBOP \cite{cheng2025achieving} & \textcolor{green!60!black}{\checkmark}Principal Statistics  & -- & -- & 82.1 & 86.7 & -- & -- & -- & -- & 81.0 & 83.9 & -- & -- & -- & -- & 71.2 & 80.0 & -- & -- \\
        OSN \cite{hu2024task}            & \textcolor{green!60!black}{\checkmark}Principal Statistics   & -- & -- &  \underline{83.6} &  \underline{88.7} & -- & -- & -- & -- & 82.6 & 84.5 & -- & -- & -- & -- & 73.2 & 80.9 & -- & -- \\
        \midrule
        \rowcolor[HTML]{EEEEEE}CoOp \cite{zhou2022learning}        & \textcolor{purple}{\ding{55}}   & 78.8   & 85.0  & 75.1 & 83.0 & 78.2  & 85.3  & 78.8 & 81.8  & 78.0 & 81.4 & 74.9 & 80.6  & 53.1 & 63.7  & 41.8 & 54.3 & 49.1 & 58.4  \\ 
        \rowcolor[HTML]{EEEEEE}CoOpOp \cite{zhou2022conditional}   & \textcolor{purple}{\ding{55}}   & 76.5   & 80.0  & 76.5 & 76.8 & 70.0  & 73.2  & 68.8 & 76.5  & 62.4 & 71.3 & 61.2 & 70.8  & 51.9 & 66.7  & 53.4 & 66.9 & 50.2 & 65.8  \\ 
       
        \rowcolor[HTML]{EEEEEE}ContinualCLIP \cite{thengane2022clip} & \textcolor{purple}{\ding{55}} & 72.8   & --    & 72.8 & --   & 72.8  & --    & 74.6 & --    & 74.6 & --   & 74.6 & --    & 60.9 & --    & 60.9 & --   & 60.9 & --    \\ 
        L2P \cite{wang2022learning}         & \textcolor{purple}{\ding{55}}   & --     & --    & 70.2 & 79.6 & --    & --    & --   & --    & 71.1 & 78.4 & --   & --    & --   & --    & 69.9 & 77.8 & --   & --   \\ 
        DualPrompt \cite{wang2022learning}  & \textcolor{green!60!black}{\checkmark}Task-aware Key-prompt   & --     & --    & 72.0 & 81.8 & --    & --    & --   & --    & 71.7 & 79.8 & --   & --    & --   & --    & 64.5 & 75.9 & --   & --   \\ 
        SLCA \cite{zhang2023slca}           & \textcolor{green!60!black}{\checkmark}Class-wise Statistics    & 81.8   & 87.3  & 80.2 & 86.6 & 79.2  & 87.8  & 82.3 & 86.0  & 80.3 & 84.4 & 80.2 & 85.4  & 75.5 & 78.9  & 73.9 & 80.0 & 71.1 & 79.4  \\  
        AttriCLIP \cite{wang2023attriclip}  & \textcolor{purple}{\ding{55}}   & 81.9   & 86.8  & 80.9 & 86.3 & 79.6  & 86.0  & 79.2 & 84.3  & 78.5 & 81.8 & 77.4 & 82.1  & 62.8 & 72.2  & 52.5 & 66.4 & 57.1 & 67.4  \\ 
        MoE-Adapter \cite{yu2024boosting}   & \textcolor{purple}{\ding{55}}  & 83.8   & 88.0  & 82.1 & 87.9 & 81.0  & 86.8  & 81.2 & 85.9  & \underline{82.9} & \underline{85.9} & \underline{82.6} & \underline{86.2}  & 78.4 & 82.5  & 75.1 & 81.4 & \underline{73.9} & \underline{80.5}  \\
        SPU \cite{zhang2024overcoming}      & \textcolor{purple}{\ding{55}}   & \underline{84.5}   & \underline{89.1}  & 82.9 &88.2 & \underline{81.2}  & 86.8  & 82.8 & 86.1  & 82.4 & 85.5 & 81.9 & 85.5  & \underline{78.8} & \underline{82.8}  & \underline{76.2} & \underline{81.8} & \underline{73.9} & 79.8  \\ 
        TaskRes-CL \cite{yu2023task}        & \textcolor{purple}{\ding{55}}   & 84.2   & 88.3  & 81.2 & 87.3 & 79.9  & 86.4  & \underline{82.9} & \underline{86.3}  & 82.5 & 85.7 & 82.1 & 85.8  & 75.4 & 78.9  & 75.0 & 80.4 & 72.7 & 78.3  \\ 
         ENGINE \cite{zhou2025external} & \textcolor{green!60!black}{\checkmark} Class-wise Statistics &  81.0  & 85.9  & 79.9 & 86.0 &  77.3 & 84.0 & -- & -- & -- & -- & -- & -- &76.3 & 82.0 & 74.0 & 81.2 & 68.9 & 76.7  \\ 
          ENGINE(w/ GDA) \cite{zhou2025external} & \textcolor{green!60!black}{\checkmark} Class-wise Statistics & 83.1   &  87.9 & 81.5 & 87.7 & 79.9  & \underline{86.9} & -- & -- & -- & -- & -- & -- &78.8 &82.5 & \textbf{78.9} &83.4   & 73.7 &  80.3 \\ 
        \rowcolor[HTML]{D7D2F5}
        DesCLIP (Ours)             & \textcolor{purple}{\ding{55}}       & \textbf{85.9} & \textbf{90.0} & \textbf{84.5} & \textbf{90.1} & \textbf{82.9} & \textbf{88.8} & \textbf{84.3} & \textbf{87.6} & \textbf{84.2} & \textbf{87.3} & \textbf{83.2} & \textbf{87.1} & \textbf{81.3} & \textbf{84.5} & 78.4 & \textbf{83.8} & \textbf{75.3} & \textbf{81.7} \\

        \midrule
        \textbf{UB}                        & \scriptsize{--} & 90.0   & --    & 90.0 & --   & 90.0  & --    & 87.6 & --    & 87.6 & --   & 87.6 & --    & 84.5 & --    & 84.5 & --   & 84.5 & --    \\
        \bottomrule[1.2pt]
    \end{tabular}

    }
\end{table*}

\textit{3) Competitors:} We compare our method against baseline and state-of-the-art methods for continual learning. These include ContinualCLIP \cite{thengane2022clip} under zero-shot conditions; non-continual adapting methods such as CoOp \cite{zhou2022learning} and CoOpOp \cite{zhou2022conditional}; and conventional continual learning method such as sequential fine-tuning (Seq FT), LwF \cite{li2017learning}, EWC \cite{kirkpatrick2017overcoming}, and recent task-orthogonal methods: OSN \cite{hu2024task} and AdaBOP \cite{cheng2025achieving}. Additionally, we evaluate VLM-specific continual learning methods, including AttriCLIP \cite{wang2023attriclip}, TaskRes-CL \cite{yu2023task}, SPU \cite{zhang2024overcoming}, MoE-Adapter \cite{yu2024boosting}, RAPF \cite{huang2025class}, and ENGINE \cite{zhou2025external}(`w/ GDA' represents using Gaussian Discriminant Analysis to correct the logits). For a broader comparison, we also incorporate techniques tailored to address continual learning in visual-only pre-trained models, such as L2P \cite{wang2022learning}, DualPrompt \cite{wang2022dualprompt}, and SLCA \cite{zhang2023slca}.

\textit{4) Implementation Details:} All experiments are conducted on two NVIDIA GeForce RTX 3090 GPUs. We adopt the pre-trained CLIP \cite{radford2021learning} ViT-L/14 and ViT-B/16 as the backbones. To ensure a fair comparison, all methods are built upon the same pre-trained CLIP backbones, and no data replay strategies are used during the continual learning process.

The model is trained for 10 epochs on the datasets for each incremental task. Stochastic Gradient Descent (SGD) \cite{bottou2010large} is employed as the optimizer, using a cosine learning rate decay schedule with a batch size of 32. The learning rate for the MLPs' weights in the CLIP's visual encoder is set to $1 \times 10^{-5}$, while the learning rate for the text shifting weights is set to 0.1. The coefficient $\alpha$ for shifting weight addition is set to 0.1. Loss balancing factors are set as $\lambda_{\text{IM}}=2.0$, $\lambda_{\text{TA}}=2.0$, and $\lambda_{\text{RIC}}=1.0$ by default. For the fine-grained CUB-200 and FGVC-Aircraft, $\lambda_{\text{IM}}$ is adjusted to 15.0, and 5.0. The threshold parameters $\delta_d$ and $\gamma$ are first estimated based on downstream tasks, and then set based on empirical observations. The specific estimation procedure and detailed analysis are presented in \ref{sec:ablation}-2).


 Additionally, we generate $n_{dsc}=30$ attribute descriptions for each class using the language assistant GPT-4 \cite{openai2023gpt4}.

\subsection{Comparison with State-of-the-art Methods}
\begin{table}[t]
    \centering
    \caption{Comparison of various methods on CIFAR100, CUB-200, and ImageNet-R ($T =10$) with CLIP ViT-B/16. \textbf{\texttt{FR.}} represents using feature replay. \textbf{\texttt{AS.}} represents using additional structure.}
    \label{tab:vit-b-16}
    \renewcommand{\arraystretch}{1.0}
    \setlength{\tabcolsep}{4pt}
    \resizebox{0.45\textwidth}{!}{%
    \fontsize{10}{12}\selectfont
     \begin{tabular}{l|cc|ccc|ccc|ccc}
        \toprule[1.2pt]
        \multicolumn{1}{c}{} & & &  
        \multicolumn{3}{c|}{CIFAR100 } & 
        \multicolumn{3}{c|}{CUB-200 } &
        \multicolumn{3}{c}{ImageNet-R } \\
        \textbf{Method}& \textbf{\texttt{FR.}} & \textbf{\texttt{AS.}} & Last & Avg & C. & Last & Avg & C. & Last & Avg & C. \\
        \midrule
        Seq FT             & \textcolor{purple}{\ding{55}} & \textcolor{purple}{\ding{55}} 
                           & 46.3 & --    & 24.2 
                           & 45.7 & --    & 39.7 
                           & 50.8 & --    & 34.3 \\ 
        \rowcolor[HTML]{EEEEEE}ContinualCLIP  
                           & \textcolor{purple}{\ding{55}} & \textcolor{purple}{\ding{55}} 
                           & 68.3 & --    & \textbf{63.6} 
                           & 55.1 & --    & \textbf{63.6} 
                           & 72.0 & --    & \textbf{63.6} \\ 
        L2P               & \textcolor{purple}{\ding{55}} & \textcolor{green!60!black}{\checkmark} 
                           & 64.5 & 72.9 & 41.8 
                           & 66.3 & 75.5 & 43.8 
                           & --   & --    & --   \\ 
        SLCA              & \textcolor{green!60!black}{\checkmark} & \textcolor{purple}{\ding{55}} 
                           & 71.5 & 79.8 & 59.1 
                           & 50.6 & 58.4 & 62.7 
                           & 72.1 & 76.2 & 57.1 \\ 
        AttriCLIP         & \textcolor{purple}{\ding{55}} & \textcolor{green!60!black}{\checkmark} 
                           & 67.0 & 77.8 & 60.3 
                           & 50.8 & 65.4 & \underline{63.1} 
                           & 70.4 & 73.7 & \underline{61.0} \\ 
        SPU               & \textcolor{purple}{\ding{55}} & \textcolor{purple}{\ding{55}} 
                           & 75.8 & 84.3 & 58.7 
                           & 67.8 & 75.6 & 59.8 
                           & 76.8 & 82.4 & 57.5 \\ 
        TaskRes-CL        & \textcolor{purple}{\ding{55}} & \textcolor{purple}{\ding{55}} 
                           & 75.6 & 83.2 & \textbf{63.6} 
                           & 67.1 & 73.8 & \textbf{63.6} 
                           & 77.4 & 81.5 & \textbf{63.6} \\ 
        MoE-Adapter       & \textcolor{purple}{\ding{55}} & \textcolor{green!60!black}{\checkmark} 
                           & 77.8 & 84.9 & \underline{62.2} 
                           & 68.3 & 76.2 & 62.9 
                           & \textbf{80.1} & 83.7 & 60.7 \\ 
        RAPF              & \textcolor{green!60!black}{\checkmark} & \textcolor{green!60!black}{\checkmark} 
                           & \underline{78.5} & \underline{85.1} & 54.0 
                           & \textbf{72.4} & \textbf{81.0} & 52.6 
                           & 76.1 & 81.5 & 54.1 \\ 

         ENGINE              & \textcolor{green!60!black}{\checkmark} & \textcolor{green!60!black}{\checkmark} 
                           & 72.3 & 82.1 & 52.8 
                           & 65.5 & 74.2 &  52.0
                           & 76.0 &79.3  & 54.6 \\ 
        ENGINE(w/ GDA)              & \textcolor{green!60!black}{\checkmark} & \textcolor{green!60!black}{\checkmark} 
                           & 75.6 & 83.7 & 52.8 
                           & \textbf{72.4} & \underline{79.5} & 52.0
                           & 76.7 & 79.7 & 54.6 \\ 
                           
        \rowcolor[HTML]{D7D2F5} DesCLIP (Ours) 
                           & \textcolor{purple}{\ding{55}} & \textcolor{purple}{\ding{55}} 
                           & \textbf{79.1} & \textbf{85.7} & 62.0 
                           & \underline{72.0} & 78.6 & 62.5 
                           & \underline{79.9} & \textbf{84.2} & 60.4 \\
        \bottomrule[1.2pt]
    \end{tabular}
    }
\end{table}

\begin{figure}[t]
  \centering
  \includegraphics[width=\linewidth]{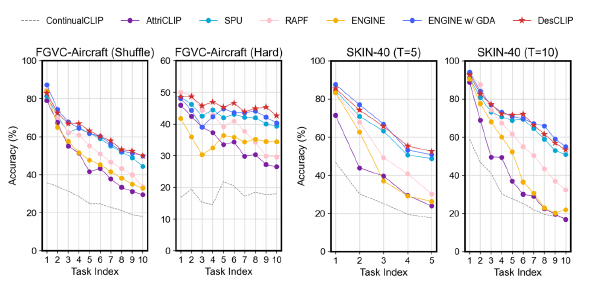}
  \vspace{-2em}
  \caption{Per-stage average accuracies on OOD datasets.}
  \label{fig:OOD}
\end{figure}

\begin{figure*}[t]
  \centering
  \includegraphics[width=0.9\linewidth]{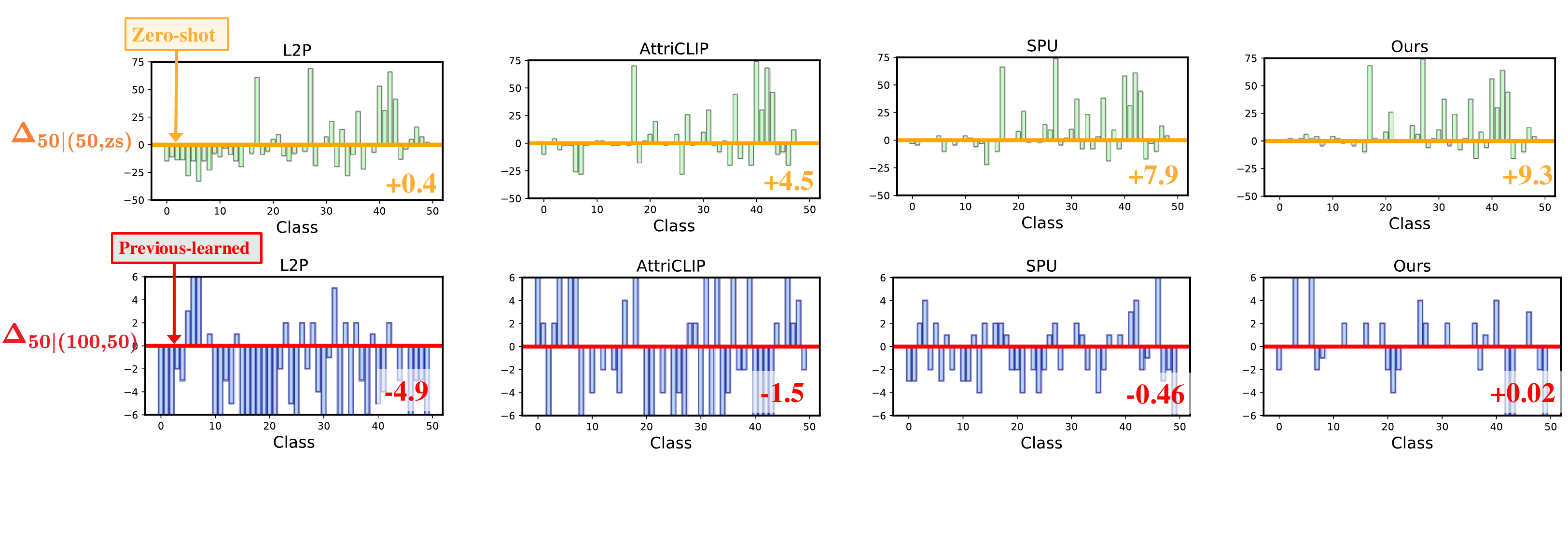}
  \vspace{-3em}
  \caption{Degradation or improvement evaluation on ImageNet-Sub with 10 incremental tasks. \textbf{Upper}: accuracy differences relative to zero-shot VLM. \textbf{Bottom}: accuracy differences relative to VLM previously learned on the first-half classes.}
  \label{fig:Deg or Imp}
\end{figure*}

The accuracy results for continual learning, including `Last' and `Avg', are summarized in Table \ref{tab:comparison}. These results are derived from comprehensive experiments conducted on multiple datasets under varying incremental task settings. 

\textit{1) Performance on Coarse Dataset:}  
CIFAR100 is selected as the coarse dataset, where VLMs generally perform well on common classes but may struggle with blurred images. As shown in Table \ref{tab:comparison}, our method consistently achieves superior performance across all incremental task stages, obtaining an `Avg' of 90.1\% under the setting of $T=10$. Notably, our approach surpasses ContinualCLIP by a significant margin of 11.6\% on the `Last' metric. Moreover, it outperforms TaskRes-CL, MoE-Adapter, SPU, and OSN by +3.3\%, +2.4\%, +1.6\%, and 0.9\%, respectively.

\textit{2) Performance on Fine-grained Datasets:}  
Table \ref{tab:comparison} demonstrates that our method consistently achieves the best performance across all incremental task stages on fine-grained datasets ImageNet-Sub and CUB-200. On ImageNet-Sub, our method outperforms TaskRes-CL \cite{yu2023task}, SPU \cite{zhang2024overcoming}, and AttriCLIP \cite{wang2023attriclip} by +1.7\%, +1.8\%, and +5.7\%, respectively. For CUB-200, our approach achieves the highest task accuracy at every stage. With $T=10$, our method surpasses SLCA by +4.5\%, SPU by +2.2\%, and TaskRes-CL by +3.4\% on the `Last' metric.

\textit{3) Performance on OOD Datasets:}
 Fig.~\ref{fig:OOD} presents the per-stage performance comparison on OOD datasets. On FGVC-Aircraft, we evaluate two settings: \textbf{\texttt{Shuffle}}, where the task class order is randomly shuffled, and \textbf{\texttt{Hard}}, where the CL model is trained and evaluated following the original class order defined in FGVC-Aircraft. Under the \textbf{\texttt{Hard}} setting, the model is required to classify visually similar but previously unseen categories within a new task, such as distinguishing between \textit{Boeing 727-200} and \textit{Boeing 737-200}. As shown in Fig.~\ref{fig:OOD}, our method consistently outperforms almost all compared approaches. Although `ENGINE w/ GDA' achieves performance comparable to ours, it relies on stored distributional information to perform Gaussian Discriminant Analysis for constructing a visual classifier's weights and bias, whereas our method does not store any information from previous tasks. In the \textbf{\texttt{Hard}} setting, our improvement is more pronounced, which gains +2.4\% `LAST' improvements compared to `ENGINE w/ GDA'. For other methods, fitting ``unfamiliar'' visual-text representations is even more detrimental to forgetting. In addition, we also report results on SKIN-40. Under both $T=5$ and $T=10$ settings, our method continues to deliver superior performance, while we observe that methods such as ENGINE and AttriCLIP almost forget all previously learned knowledge, with their `LAST' scores close to or even lower than the zero-shot baseline of ContinualCLIP. These results demonstrate the OOD generalization capability of our method. It should be emphasized that this performance gain relies on proper GA descriptions to bridge the matching gap between visual and textual modalities, which constitutes the prerequisite for the effectiveness of our proposed \textbf{AEF} mechanism.
 
 
\begin{figure}[t]
  \centering
  \includegraphics[width=0.9\linewidth]{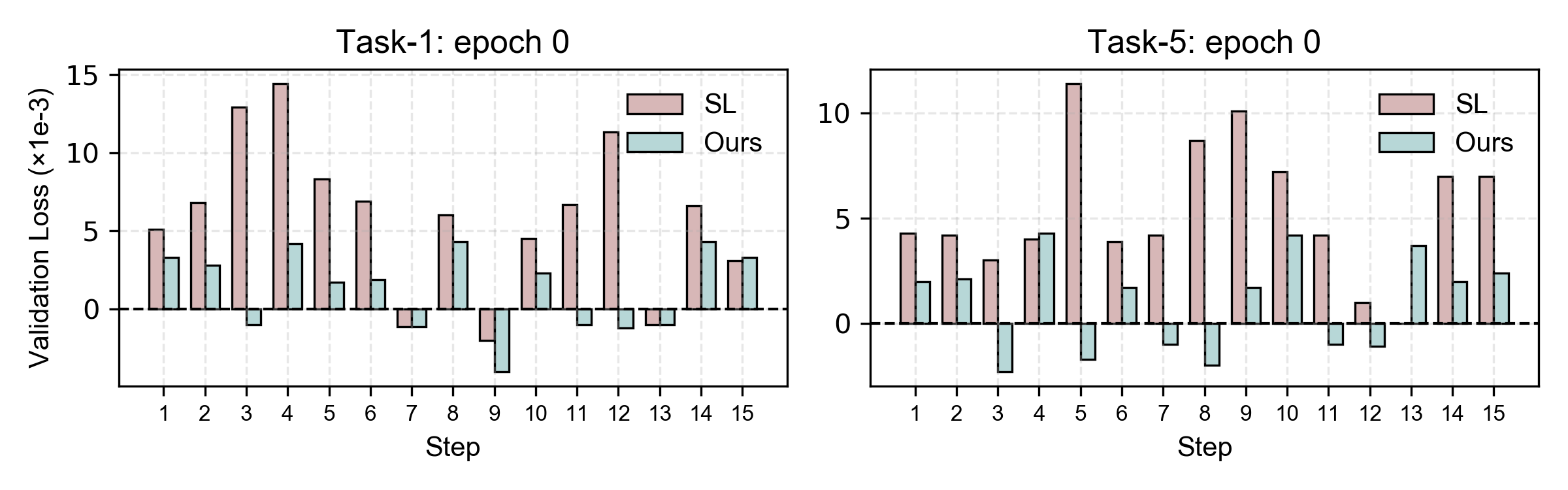}
  \vspace{-1.2em}
  \caption{Loss variation of each steps on validation set of ImageNet-1K.}
  \label{fig:val_loss_2tasks}
\end{figure}

\begin{table}[t]
\centering
\setlength{\tabcolsep}{10pt} 
\renewcommand{\arraystretch}{1} 
\caption{Ablation study of each component on CUB-200.}
\label{tab:ablation1}
\resizebox{0.5\textwidth}{!}{ 
\begin{tabular}{c|ccc|cc|cc}
\toprule[1.2pt]
\multirow{2}{*}{\textbf{Settings}} & \multirow{2}{*}{\textbf{RIC}} & \multirow{2}{*}{\textbf{IM}} & \multirow{2}{*}{\textbf{TA}} & \multicolumn{2}{c|}{$T=10$} & \multicolumn{2}{c}{$T=20$}  \\
 & & & & $\Delta$Last& $\Delta$Avg& $\Delta$Last& $\Delta$Avg\\
\midrule
Zero-shot & -& -& - & 60.9& 66.0& 60.9& 66.0\\
\rowcolor[HTML]{EEEEEE} V.E.(full)& \checkmark &  & & \textcolor{blue}{-7.8} & \textcolor{blue}{-7.3} & \textcolor{blue}{-8.2} & \textcolor{blue}{-7.9}  \\
\rowcolor[HTML]{EEEEEE} V.E.& \checkmark &  & & \textcolor{blue}{-2.1} & \textcolor{blue}{-2.1} & \textcolor{blue}{-2.7} & \textcolor{blue}{-2.6}   \\
V.E. & \checkmark & \checkmark & & +3.6 & +3.9 & +2.0 & +2.2   \\
V.E.+\textbf{TEC} &\checkmark & \checkmark & & +12.5 & +13.1 & +11.4 & +12.0  \\
only \textbf{TEC} &\checkmark &  & & +14.1 & +14.4 & +11.8 & +12.3  \\
\rowcolor[HTML]{D7D2F5}V.E.+\textbf{TEC} & \checkmark & \checkmark & \checkmark & \textbf{+17.5}& \textbf{+17.8}& \textbf{+14.4} & \textbf{+15.7} \\
\bottomrule[1.2pt]
\end{tabular}
}
\end{table}

\begin{table}[t]
\centering
\setlength{\tabcolsep}{11pt}      
\renewcommand{\arraystretch}{1}   
\caption{Settings of threshold parameters $\delta_d$ and $\gamma$. The estimated values are highlighted by {\color[HTML]{D7D2F5}\rule{10pt}{6pt}}.}
\label{tab:thres}
\resizebox{0.5\textwidth}{!}{     
\begin{tabular}{lcccccccc}
\toprule[1.2pt]
\multirow{3}{*}{\textbf{Dataset}}
  & \multicolumn{4}{c}{CLIP ViT-L/14}
  & \multicolumn{4}{c}{CLIP ViT-B/16} \\
\cmidrule(lr){2-5} \cmidrule(lr){6-9}
  & \multicolumn{2}{c}{$\delta_d$}
  & \multicolumn{2}{c}{$\gamma(e^{-2})$}
  & \multicolumn{2}{c}{$\delta_d$}
  & \multicolumn{2}{c}{$\gamma(e^{-2})$} \\
\cmidrule(lr){2-3} \cmidrule(lr){4-5} \cmidrule(lr){6-7} \cmidrule(lr){8-9}

CIFAR100       & 0.2  & \cellcolor[HTML]{D7D2F5}{0.19} & 1.5 & \cellcolor[HTML]{D7D2F5}{1.3} & 0.2  & \cellcolor[HTML]{D7D2F5}{0.21} & 1.5 & \cellcolor[HTML]{D7D2F5}{1.7} \\
ImageNet-Sub   & 0.25  & \cellcolor[HTML]{D7D2F5}{0.21}  & 3 & \cellcolor[HTML]{D7D2F5}{2.8}  & --  & --  & -- & -- \\
CUB-200        & 0.25  & \cellcolor[HTML]{D7D2F5}{0.24}  & 1.5 & \cellcolor[HTML]{D7D2F5}{1.7}  & 0.3 & \cellcolor[HTML]{D7D2F5}{0.30}  & 1.5 & \cellcolor[HTML]{D7D2F5}{1.8} \\
ImageNet-R     & 0.2 & \cellcolor[HTML]{D7D2F5}{0.19}  & 1.5 & \cellcolor[HTML]{D7D2F5}{1.8}  & 0.25  & \cellcolor[HTML]{D7D2F5}{0.25}  & 3   & \cellcolor[HTML]{D7D2F5}{2.2} \\
FGVC-Aircraft  & 0.2  & \cellcolor[HTML]{D7D2F5}{0.22}  & 3   & \cellcolor[HTML]{D7D2F5}{2.0}  & 0.25 & \cellcolor[HTML]{D7D2F5}{0.27}  & 3   & \cellcolor[HTML]{D7D2F5}{2.0} \\
SKIN-40        & 0.2  & \cellcolor[HTML]{D7D2F5}{0.22}  & 3   & \cellcolor[HTML]{D7D2F5}{2.2}  & 0.25 & \cellcolor[HTML]{D7D2F5}{0.26}  & 3   & \cellcolor[HTML]{D7D2F5}{2.4} \\
\bottomrule[1.2pt]
\end{tabular}
}
\end{table}

\textit{4) Performance on CLIP ViT-B/16 Backbone:}  
Table \ref{tab:vit-b-16} presents the continual learning performance across 10 tasks utilizing CLIP ViT-B/16 backbone \cite{radford2021learning}. Without relying on replay mechanisms or incorporating additional architectural components, our method almost outperforms state-of-the-art approaches on CIFAR100, CUB-200, and ImageNet-R. Notably, while achieving outstanding CL performance, our method maintains accuracies over 60$\%$ on ImageNet-1K (C.), closely matching the original pretrained CLIP accuracy of 63.6$\%$. In contrast, methods such as RAPF and ENGINE, although leveraging highly specialized adapter architectures to boost downstream CL performance, suffer from a noticeable decline in general knowledge recognition ability, as reflected in the significant drop of C.

\textit{5) Zero-shot Degradation or Improvement?}  
We propose a metric to evaluate the relative accuracy changes for all classes with respect to the zero-shot CLIP model after incremental tasks. Specifically, we calculate the class-wise accuracy difference between the zero-shot CLIP and the trained CLIP model after half of the incremental tasks (50 classes), denoted as $\Delta_{50|(50,zs)}$, and present the results in Fig. \ref{fig:Deg or Imp} (Upper). Our method achieves the highest averaged $\Delta_{50|(50,zs)}$, reaching +9.3. This demonstrates that our approach experiences the least zero-shot degradation while achieving superior performance, attributed to the more robust visual-textual relationships established during continual learning compared to the original vision-class text relationships. As shown in Fig. \ref{fig:val_loss_2tasks}, during training, our method maintains a lower validation loss on the balanced subset of ImageNet-1K compared to the standard slow leaner (SL) \cite{zhang2023slca}, thereby mitigating the risk of forgetting pretrained knowledge.

Furthermore, we calculate the accuracy difference between the `after all-class incremental learning' state and the `only learning on the first-50 classes' state, denoted as $\Delta_{50|(100,50)}$, as shown in Fig. \ref{fig:Deg or Imp} (Bottom). Our method achieves an averaged $\Delta_{50|(100,50)}$ of +0.02, indicating effective knowledge transfer from previous tasks. This improvement can be attributed to the establishment of a strong connection between general visual-textual knowledge during continual learning.  

\begin{figure}[t]
  \centering
  \includegraphics[width=0.7\linewidth]{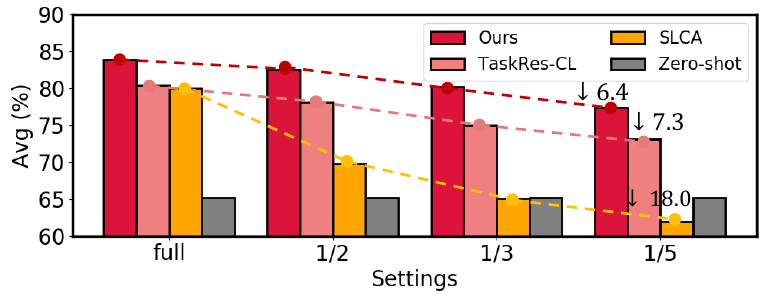}
  \vspace{-0.5em}
  \caption{Performance comparison in few-shot case.}
  \label{fig:few-shot}
\end{figure}

\begin{figure}[t]
    \centering
    \includegraphics[width=0.9\linewidth]
    {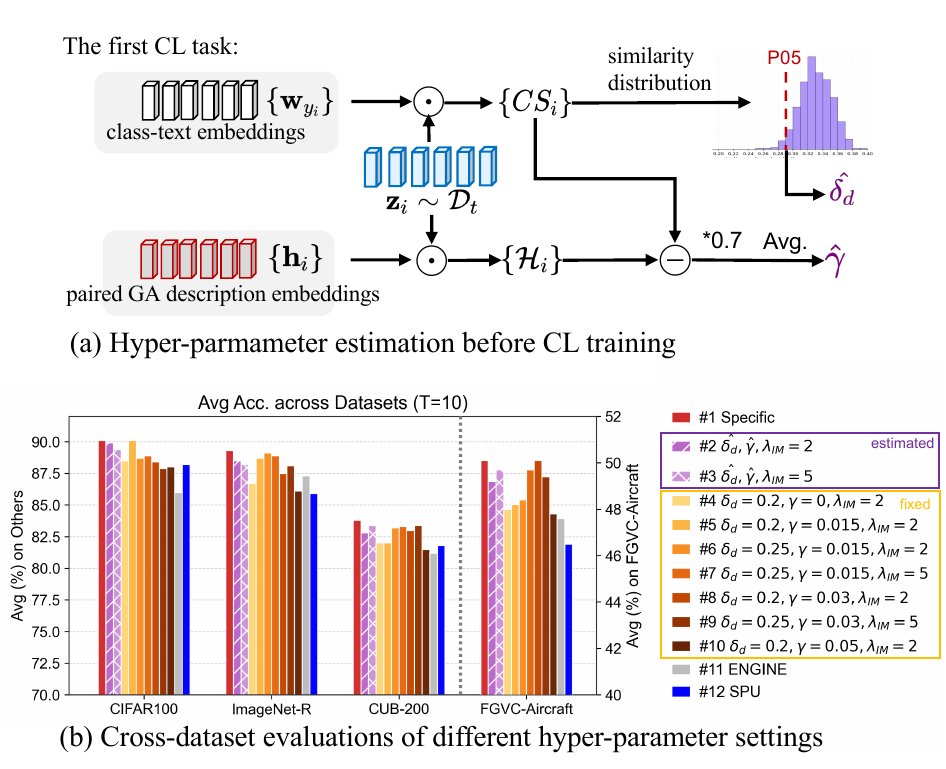}
    \vspace{-0.5em}
    \caption{Hyperparameter estimation and cross-dataset evaluations.}
    \label{fig:Hyperparam_cross_dataset}
\end{figure}

 \begin{figure}[t]
  \centering
  \includegraphics[width=\linewidth]{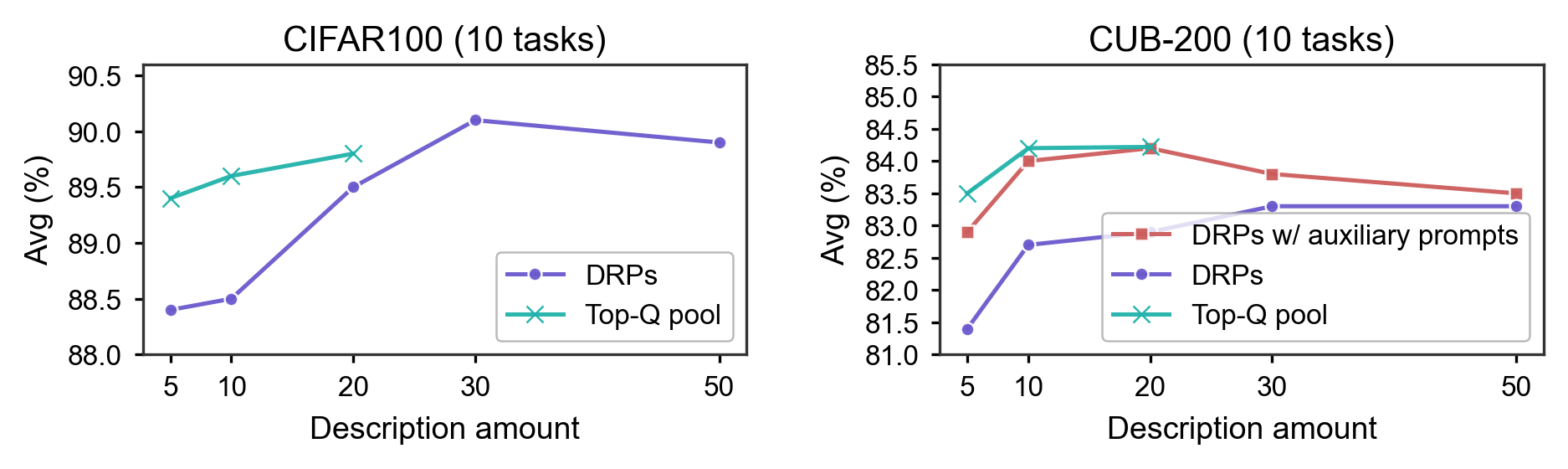}
  \vspace{-1.5em}
  \caption{Ablation study of generated GA description amount.}
  \label{fig:ablation_des_nums}
\end{figure}

\textit{6) Few-shot Performance:}  
Fig. \ref{fig:few-shot} illustrates the continual learning performance in few-shot scenarios on the CUB-200 dataset. When the number of training samples per task is reduced from `full' to `1/5', our method maintains a competitive advantage, with an average performance drop of only -6.4. This decline is significantly smaller compared to TaskRes-CL's -7.3 and SLCA's -18.0, showcasing the robustness of our approach under challenging conditions.

\subsection{Ablation Study}
\label{sec:ablation}
\textit{1) Effectiveness of Each Component:}  
We conduct detailed ablation studies on various components of our DesCLIP framework. As shown in Table \ref{tab:ablation1}, the zero-shot model is used as the baseline, and the relative improvements or declines are reported. It is evident that a fully fine-tuned visual encoder (V.E.(full)) suffers from catastrophic forgetting. Without data replay, merely fine-tuning the visual encoder (V.E.) partially (tuning the first MLPs in each Transformer block) proves inadequate for continual learning \cite{zhang2024overcoming}. Based on V.E., by integrating instance matching (\textbf{IM}) with visually filtered attribute description embeddings, we achieve a relative improvement of 2$\sim$3\% over the zero-shot baseline. This improvement highlights the effectiveness of \textbf{IM} in mitigating forgetting, which typically arises from overfitting specific downstream task classes. Furthermore, the impact of text embedding calibration (\textbf{TEC}) is notable. The combination of \text{V.E.}+\textbf{TEC}+\textbf{IM}+\textbf{TA}, leveraging the filtered attribute description embeddings, yields the best performance, showcasing the synergy between these components.

\textit{2) Hyperparameter Analysis:}
The hyperparameter setting is crucial for the effectiveness of the proposed \textbf{AEF} mechanism. Given the inherent differences in visual–textual similarity across downstream tasks, it is critical to efficiently determine suitable hyperparameters for CL. In our study, we first obtain preliminary estimates of the threshold parameters $\delta_d$ and $\gamma$, and then refine their values through empirical tuning to ensure better performance across tasks. Specifically, the estimation process is as illustrated in Fig.~\ref{fig:Hyperparam_cross_dataset}(a). In the first CL task, we compute the \textit{similarity distribution} between training images and their corresponding class-text embeddings, and take the 5th percentile (P05) of this distribution as the estimated value $\hat{\delta_d}$. Meanwhile, for each visual instance, we compute the similarity with its most aligned GA description embedding, \(
\mathcal{H}_{i}= \left \langle \frac{\mathbf{z}_i}{\|\mathbf{z}_i\|}, \frac{\mathbf{h}_i}{\|\mathbf{h}_i\|} \right \rangle,\)
and estimate $\hat{\gamma}$ via \(
\hat{\gamma} = \mathbb{E}_{i=1:N_{t=1}} \left(\mathcal{H}_i - CS_i\right) \cdot 0.7,\) which provides a reasonable initialization and is further adjusted in practice to mitigate the effect of noise on the \textbf{AEF} mechanism in downstream tasks.
The downstream-task settings of specific threshold parameters and estimated parameters (highlighted) are reported in TABLE. \ref{tab:thres}.

We analyze the robustness of hyperparameters obtained through this unified estimation strategy and compare the results with those trained under dataset-specific hyperparameters. In addition, we provide further comparisons to show that, even with a unified hyperparameter setting strategy, our method still surpasses competing approaches. The CL performance across multiple datasets is illustrated in Fig.~\ref{fig:Hyperparam_cross_dataset}(b). In particular, we evaluate the specific configuration for each dataset (\#1), the estimated threshold hyperparameters (\#2–3), different fixed hyperparameters (\#4–10), and two recent CL methods (\#11–12). These results demonstrate that our approach exhibits low dependence to hyperparameter selection, since both estimated and fixed settings still achieve performance that is almost consistently superior to recent CL baselines. Moreover, we observe that fine-grained downstream tasks, such as CUB-200 and FGVC-Aircraft, generally favor higher values of $\lambda_{\textbf{IM}}$. However, even with a fixed loss coefficient $\lambda_{\textbf{IM}}$, competitive CL performance can still be achieved. On the other hand, filtering of GA description embeddings proves to be a key factor: overly conservative filtering (e.g., $\gamma=0$) or excessively strict filtering (e.g., $\gamma=0.05$) leads to a noticeable drop in performance.


\begin{figure*}[t]
  \centering
  \includegraphics[width=\linewidth]{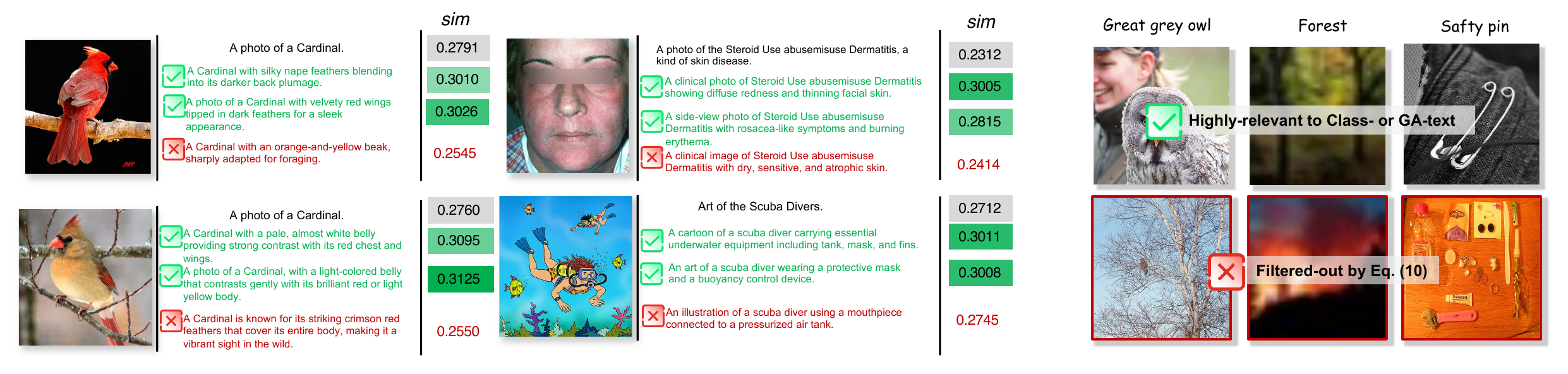}
  \vspace{-1.5em}
  \caption{Visualizations of \textbf{Left:} compliance and noncompliance GA descriptions decided by Eq.~\ref{eq:AEF}; \textbf{Right:} Filtered-out visual instances by Eq.~\ref{eq_delta_d}}.
  \label{fig:visual1}
\end{figure*}

\textit{3) Generated Attribute Descriptions:} We analyze the impact of the generated description amount on the performance of continual learning, as shown in Fig. \ref{fig:ablation_des_nums}. For coarse CIFAR100, we only employed the auxiliary prompt ``\texttt{Please maintain diversity between descriptions}" to generate $n_{dsc}$ attribute descriptions for each class. We find that the model performed optimally when $n_{dsc}$ is set to 30. In contrast, for the fine-grained CUB-200, we have discovered that incorporating additional auxiliary prompts, ``\texttt{Describe a/an <class~name>'s attributes from its beak, eyes, body, belly, tail, wings, breast, etc.}" results in improved performance (red star in Fig. \ref{fig:ablation_des_nums}) even with a smaller $n_{dsc}$. This approach enhances the quality of the descriptions, provided that the language assistant possesses sufficient knowledge of the object's details. 

Moreover, redundancy was found among the generated GA descriptions. In CIFAR100, for instance, only a limited number of GA embedding candidates were selected as highly relevant to the corresponding visual instances for each class. To further optimize the GA candidate pool, we explored a static pre-filtering strategy before training. Specifically, we applied a fixed AEF-based selection to evaluate the matching confidence of each GA embedding without updating the CLIP model. For each class, GA candidates were ranked by their selection frequency, and the top-$Q$ candidates were retained to form a reduced GA pool. As shown in Fig.~\ref{fig:ablation_des_nums}, this pre-filtering achieves a modest yet consistent improvement in CL performance, particularly on CIFAR100. Notably, using only 10 pre-filtered GA descriptions surpasses the performance achieved with 20 unfiltered candidates. In contrast, on CUB-200, where GA descriptions are inherently more distinctive and less redundant, the impact of pre-filtering is less significant. These results suggest that while GA diversity is crucial, a compact set of high-confidence descriptions is sufficient to guide instance matching effectively.

 \begin{figure}[t]
  \centering
  \includegraphics[width=\linewidth]{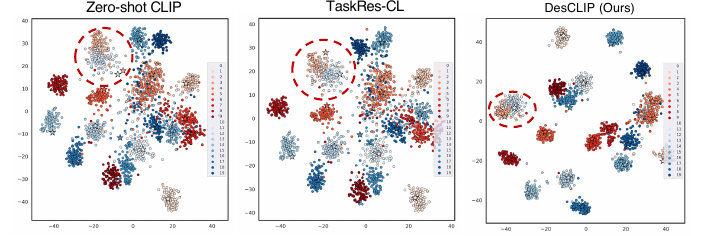}
  \vspace{-1em}
  \caption{t-SNE visualizations on CIFAR100. Dots: visual representations; Stars: text embeddings.}
  \label{fig:visual4}
\end{figure}

\subsection{Visualizations}

\textit{1) Description of Compliance and Noncompliance:} In Fig.~\ref{fig:visual1}(\textbf{Left}), we present visual instances alongside GA descriptions that either match or do not match the attribute features filtered by \textbf{AEF}, together with their corresponding similarity scores. This illustrates the degree of visual-text alignment for each GA under both compliant and non-compliant conditions. Compared to manually designed prompts, the GA descriptions filtered by \textbf{AEF} exhibit significantly stronger visual relevance, especially in specialized domains such as dermatological diagnosis.

\textit{2) Filtered-out Visual Instances:} As shown in Fig.~\ref{fig:visual1}(\textbf{Right}), it can be observed that, compared to the retained visual instances, the filtered visual instances lack relevance to both general attributes and class-specific information.

\textit{3) t-SNE Visualization:} Fig.~\ref{fig:visual4} presents the t-SNE visualizations of CLIP's visual representations and text embeddings of each class. We employed distribution alignment to transform text embeddings into the visual representation space, rather than using the original text embedding space. Intuitively, compared to Zero-shot, TaskRes-CL \cite{yu2023task} adjusts text embeddings in downstream incremental tasks without optimizing visual representations; SPU \cite{zhang2024overcoming} optimizes visual representations but struggles to align visual-text representations for unfamiliar classes. In contrast, our method achieves superior alignment of visual-text representations in incremental tasks.

\begin{table}[t]
\centering
\setlength{\tabcolsep}{12.0pt} 
\renewcommand{\arraystretch}{1} 
\caption{Comparison of different description sources and their CL performances on CUB-200.}
\label{tab:description_quality}
\resizebox{0.5\textwidth}{!}{ 
\begin{tabular}{l|c|c|c}
\toprule[1.2pt]
\textbf{Description Types} & \textbf{LLM Types}& \textbf{Num. / Class} & Avg (\%) \\
\midrule
\rowcolor[HTML]{EEEEEE}CUB-attributes$^1$ \cite{reed2016learning}& - & 300 & 67.8 \\
\rowcolor[HTML]{EEEEEE}CUB-attributes$^2$ \cite{reed2016learning}& - & 30 & 65.5 \\
DRPs & GPT-3 & 30 & 81.0 \\
DRPs & GPT-4 & 30 & 82.9 \\
DRPs & DeepSeek-R1 & 30 & 82.2 \\
DRPs w/ auxiliary prompts & DeepSeek-R1 & 30 & 83.2 \\
\rowcolor[HTML]{D7D2F5} DRPs w/ auxiliary prompts & GPT-4 & 30 & \textbf{83.8} \\
\bottomrule[1.2pt]
\end{tabular}
}
\label{tab:different description sources}
\end{table}

\begin{figure*}[t]
  \centering
  \includegraphics[width=0.9\linewidth]{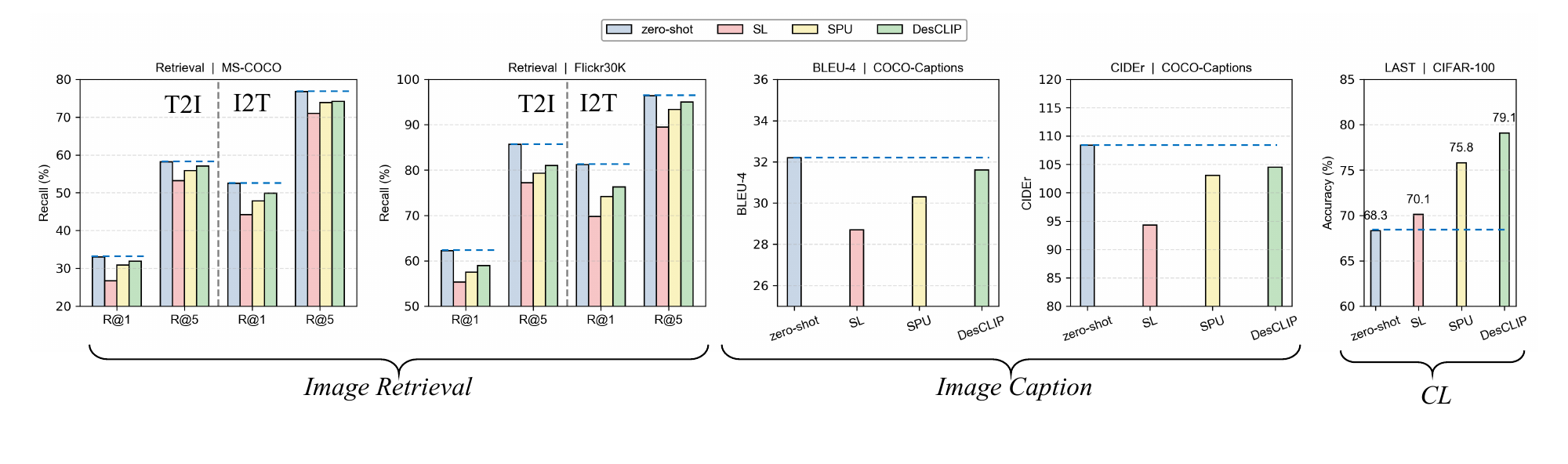}
  \vspace{-2em}
  \caption{Cross-task validations on Image Retrieval and Image Caption with CLIP ViT-B/16.}
  \label{fig:combined-retrieval-cap}
\end{figure*}

\section{Discussions}
\subsection{In-depth Study of Text Generation Strategies}
\textit{1) GA Description Source:} We conducted an in-depth study on the impact of GA descriptions from different sources on continual learning performance using the CUB-200 dataset. As shown in Table~\ref{tab:different description sources}, we compared several types of descriptions: CUB-attributes curated by Reed et al. \cite{reed2016learning}, GA descriptions generated by GPT-3 \cite{brown2020language}, DeepSeek-R1 \cite{guo2025deepseek}, and GPT-4 \cite{openai2023gpt4}. Our results indicate that the CUB-attributes, which focus exclusively on object-level attributes and omit explicit class identifiers, result in inferior performance. This outcome suggests that CLIP also benefits from class-level semantic cues, which are absent in such formulations. In contrast, GA descriptions produced by other LLMs, such as DeepSeek-R1 \cite{guo2025deepseek}, achieve comparable continual learning performance, especially when auxiliary prompts are employed to guide the model in emphasizing key visual attributes.

\textit{2) Comparison of LLM-based Description Generation Method:}
Table~\ref{tab:last_metric_llm} reports the performance gains obtained by replacing handcrafted templates with LLM-generated GA descriptions. Zero-shot CLIP with handcrafted prompts serves as the baseline. We compare several representative LLM-based text generation methods, including CuPL \cite{pratt2023does}, DCLIP \cite{menon2022visual}, and advanced strategies such as FuDD \cite{esfandiarpoor2023follow} and MPVR \cite{mirza2024meta}, with the following variants:
\begin{itemize}
    \item \textbf{FuDD\textsuperscript{AS}}: Differential descriptions generated using \emph{all seen classes}.
    \item \textbf{FuDD\textsuperscript{CL}}: Differential descriptions generated only from \emph{classes observed in the current task}.
    \item \textbf{MPVR\textsuperscript{G}}: Rich semantic prompts generated via meta-prompting using GPT-3.5.
    \item \textbf{MPVR\textsuperscript{M}}: Rich semantic prompts generated via meta-prompting using Mixtral-7B~\cite{jiang2024mixtral}.
\end{itemize}
We observe that replacing handcrafted templates with these LLM-based textual descriptions consistently improves performance, both in zero-shot settings and under CL methods. However, all variants remain inferior to our proposed DesCLIP. Furthermore, Table~\ref{tab:pool_construction} presents results where the generated texts from these representative strategies are utilized to construct the GA description candidate pool, essentially functioning as variants of DesCLIP. Based on our observations, advanced LLM-based methods like MPVR tend to produce more detailed and comprehensive semantic descriptions, while FuDD emphasizes generating discriminative attributes to resolve inter-class ambiguities. Compared to simpler generation methods such as CuPL and DCLIP, these advanced approaches yield significantly better performance in CL scenarios.

\begin{table}[t]
\centering
\setlength{\tabcolsep}{8pt}
\renewcommand{\arraystretch}{1}
\caption{`Last' ($T{=}10$) improvement compared to zero-shot CLIP ($\%$). `Handcraft' represents using the template `A photo of a $<class~name>$.' to obtain class-text embeddings. The values in parentheses indicate the relative gain compared to using handcraft template.}
\label{tab:last_metric_llm}
\resizebox{0.5\textwidth}{!}{
\begin{tabular}{l|l|c|c|c}
\toprule[1.2pt]
 & \textbf{Method} & \textbf{CL Training} & ImageNet-Sub & CUB-200 \\
\midrule
\multirow{1}{*}{} 

\multirow{6}{*}{\makecell[l]{LLM-generated \\ texts}}
& CuPL & \textcolor{purple}{\ding{55}} & +1.1 & +0.8 \\
& DCLIP & \textcolor{purple}{\ding{55}} & +3.8 & +1.2 \\
& FuDD$^{CL}$ & \textcolor{purple}{\ding{55}} & +1.7 & +1.6 \\
& FuDD$^{AS}$ & \textcolor{purple}{\ding{55}} & +4.1 & +3.6 \\
& MPVR$^G$ & \textcolor{purple}{\ding{55}} & +3.6 & +3.0 \\
& MPVR$^M$ & \textcolor{purple}{\ding{55}} & +3.3 & +2.5 \\
\midrule
\multirow{3}{*}{\makecell[l]{CL methods \\ + Handcraft}}
& AttriCLIP & \textcolor{green!60!black}{\checkmark} & +3.9 & -8.4 \\
& TaskRes-CL & \textcolor{green!60!black}{\checkmark} & +7.9 & +14.1 \\
& SPU & \textcolor{green!60!black}{\checkmark} & +7.8 & +15.3 \\
\midrule
\multirow{6}{*}{Combined} 
& TaskRes-CL + CuPL & \textcolor{green!60!black}{\checkmark} 
& +7.9 (+0.0) & +14.5 (+0.4) \\
& TaskRes-CL + FuDD$^{CL}$ & \textcolor{green!60!black}{\checkmark} 
&+8.5 (+0.6) &+15.9 (+1.8) \\
& TaskRes-CL + MPVR$^G$ & \textcolor{green!60!black}{\checkmark} 
& +7.5 (\textcolor{blue}{-0.4}) & +15.3 (+1.2) \\
& SPU + CuPL & \textcolor{green!60!black}{\checkmark} 
& +7.1 (\textcolor{blue}{-0.7}) & +15.8 (+0.5) \\
& SPU + FuDD$^{CL}$ & \textcolor{green!60!black}{\checkmark} 
& +8.0 (+0.2) & +16.3 (+1.0) \\
& SPU + MPVR$^G$ & \textcolor{green!60!black}{\checkmark} 
& +7.1 (\textcolor{blue}{-0.7}) & +16.1 (+0.8) \\

\rowcolor[HTML]{D7D2F5} & DesCLIP & \textcolor{green!60!black}{\checkmark} & \textbf{+9.6} & \textbf{+17.5} \\
\bottomrule[1.2pt]
\end{tabular}
}
\end{table}

\begin{table}[t]
\centering
\setlength{\tabcolsep}{14pt}
\renewcommand{\arraystretch}{1}
\caption{Comparison of different GA description candidate pool construction methods and pool sizes. We report `Avg' and `Last' ($T=10$). `DRP$^{*}$' represents using additional auxiliary prompts.}
\label{tab:pool_construction}
\resizebox{0.5\textwidth}{!}{
\begin{tabular}{l|c|cc|cc}
\toprule[1.2pt]
\makecell[c]{\textbf{GA candidate pool} \\ \textbf{Construction}} & \textbf{Pool size} & \multicolumn{2}{c|}{ImageNet-Sub} & \multicolumn{2}{c}{CUB-200} \\
& & Last & Avg & Last  & Avg \\
\midrule
+ DRP & 10 & 82.7 & 86.1 & 77.9 & 82.7 \\
\rowcolor[HTML]{D7D2F5}+ DRP & 30 & 84.2 & 87.3 & 77.6 &  82.5 \\
\rowcolor[HTML]{D7D2F5}+ DRP$^{*}$ & 30 & -- & -- & 79.0 &  83.3 \\
\rowcolor[HTML]{D7D2F5}+ DRP$^{*}$ & 20 & -- & -- & \textbf{80.1} & \textbf{84.2} \\
\midrule
+ CuPL & 30 & 82.2 & 84.9 & 76.1 & 80.3 \\
+ DCLIP & 30 & 83.8 & 87.0 & 78.5 & 82.9 \\
+ FuDD$^{CL}$ & -- & \textbf{85.4} & \textbf{88.6} & 77.1 & 81.2 \\
+ MPVR$^G$ & 30 & 83.4 & 85.8 & 77.9 & 83.0 \\
+ MPVR$^G$ & 100 & 85.2 & 88.1 & 79.9 & 83.2 \\
\bottomrule[1.2pt]
\end{tabular}
}
\end{table}

\subsection{Preservation of Cross-Task General Knowledge}
Pre-trained VLMs are applicable not only to image recognition but also to a broad spectrum of vision–language tasks. Therefore, under the CL setting, it is crucial to examine whether \emph{cross-task general knowledge} can be effectively preserved. To this end, we evaluate performance drift on two representative downstream tasks. For image retrieval, we utilize the CL-adapted CLIP weights and measure both \emph{image-to-text} (I2T) and \emph{text-to-image} (T2I) retrieval performance on the MS-COCO \cite{lin2014microsoft} and Flickr30K \cite{young2014image} benchmarks. For image captioning, we adopt the ClipCap \cite{mokady2021clipcap} framework by reusing the continually adapted CLIP weights and assess caption quality on the COCO-Captions \cite{chen2015microsoft} dataset, reporting BLEU-4 \cite{papineni2002bleu} and CIDEr \cite{vedantam2015cider} scores. As illustrated in Fig.\ref{fig:combined-retrieval-cap}, DesCLIP not only achieves superior performance on the CL tasks, but also maintains competitive performance on the image retrieval task. Compared to SL and SPU, it exhibits a smaller drop in both R@1 and R@5 scores, remaining close to the original zero-shot CLIP performance. Similarly, DesCLIP demonstrates the least degradation in image captioning quality, indicating its strong ability to preserve task-agnostic general knowledge throughout the continual learning process.

\subsection{Limitations}
Observed in experimental trials, the effectiveness of our DesCLIP hinges on the language assistant (or real experts) being knowledgeable about the general attribute features of objects. This can pose challenges in certain applications that require indirect reasoning, where it is difficult to accurately describe the representative features of a vehicle model associated with a specific license plate. Furthermore, our method may face limitations when expert knowledge is unavailable or when encountering OOD visual objects that are inherently difficult to describe. In such scenarios, approaches that mine patterns solely from single-modal visual representations (e.g., GDA) could be more effective, as they bypass the dependency on precise attribute descriptions and rely instead on direct visual regularities.

\section{Conclusions}
Current research on VLM-based continual learning primarily focuses on acquiring downstream knowledge through opaque parameter updates and implicit model adjustments, often overlooking the structured relationship between general and task-specific knowledge. In this paper, we propose DesCLIP,  a framework that harnesses general attribute (GA) descriptions to enhance VLMs in establishing robust \textit{vision-GA-class text} associations. By going beyond the traditional connections between visual inputs and class texts, DesCLIP employs a language assistant to generate candidates for attribute descriptions through tailored prompting. Additionally, we implement an anchor-based embedding filter (\textbf{AEF}) to extract highly relevant description embeddings, which serve as paired text embeddings for instance matching (\textbf{IM}). Addtionally, we perform text embedding calibration (\textbf{TEC}) which allows for the progressive calibration of rudimentary text embeddings to align with representative GA representations. Our extensive experiments validate the effectiveness and advancements of DesCLIP, demonstrating its superior performance over existing pretrained and VLM-based continual learning methods.

\bibliographystyle{IEEEtran}


\bibliography{IEEEabrv,DesCLIP}

\begin{thebibliography}{1}

\bibitem{ams}
{\it{Mathematics into Type}}, American Mathematical Society. Online available:

\bibitem{oxford}
T.W. Chaundy, P.R. Barrett and C. Batey, {\it{The Printing of Mathematics}}, Oxford University Press. London, 1954.

\bibitem{lacomp}{\it{The \LaTeX Companion}}, by F. Mittelbach and M. Goossens

\bibitem{mmt}{\it{More Math into LaTeX}}, by G. Gr\"atzer

\bibitem{amstyle}{\it{AMS-StyleGuide-online.pdf,}} published by the American Mathematical Society

\bibitem{Sira3}
H. Sira-Ramirez.   On the sliding mode control of nonlinear systems,'' \textit{Systems \& Control Letters}, vol. 19, pp. 303--312, 1992.

\bibitem{Levant}
A. Levant.   Exact differentiation of signals with unbounded higher derivatives,''  in \textit{Proceedings of the 45th IEEE Conference on Decision and Control}, San Diego, California, USA, pp. 5585--5590, 2006.

\bibitem{Cedric}
M. Fliess, C. Join, and H. Sira-Ramirez.   Non-linear estimation is easy,'' \textit{International Journal of Modelling, Identification and Control}, vol. 4, no. 1, pp. 12--27, 2008.

\bibitem{Ortega}
R. Ortega, A. Astolfi, G. Bastin, and H. Rodriguez.   Stabilization of food-chain systems using a port-controlled Hamiltonian description,'' in \textit{Proceedings of the American Control Conference}, Chicago, Illinois, USA, pp. 2245--2249, 2000.

\end{thebibliography}

\begin{IEEEbiography}[{\includegraphics[width=1in,height=1.25in,clip,keepaspectratio]{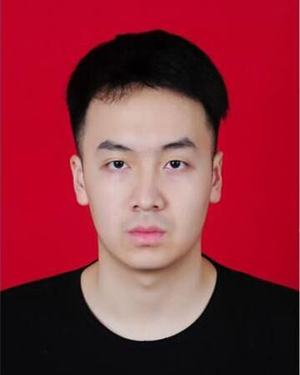}}]{Chiyuan He} recieved his M.S. degree in Information and Communication Engineering at the University of Electronic Science and Technology of China (UESTC). He is currently pursuing the Ph.D. degree in UESTC. His main research interests include lifelong learning and multimodal intelligence.
\end{IEEEbiography}

\begin{IEEEbiography}
[{\includegraphics[width=1in,height=1.25in,clip,keepaspectratio]{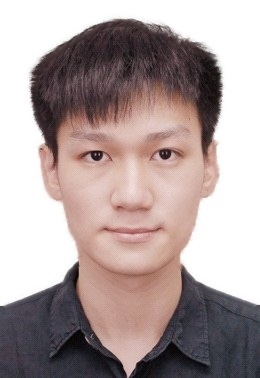}}]{Zihuan Qiu} is currently pursuing the Ph.D. degree in University of Electronic Science and Technology of China, Chengdu, China. His current research interests include continual learning and machine learning.
\end{IEEEbiography}

\begin{IEEEbiography}[{\includegraphics[width=1in,height=1.25in,clip,keepaspectratio]{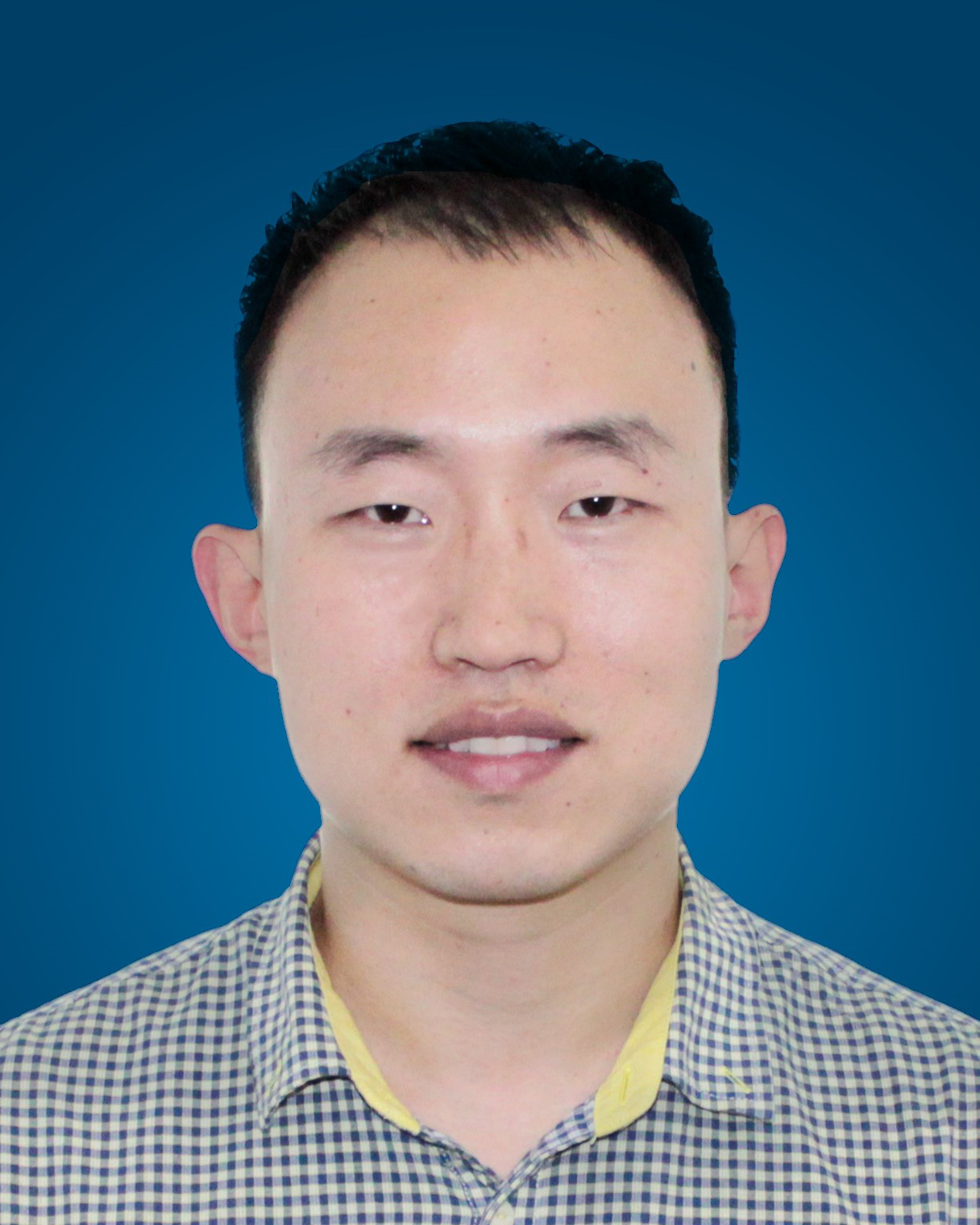}}]{Fanman Meng} (Member, IEEE) received the Ph.D. degree in Signal and Information Processing from the University of Electronic Science and Technology of China, Chengdu, China, in 2014. From 2013 to 2014, he was a Research Assistant with the Division of Visual and Interactive Computing, Nanyang Technological University, Singapore. He is currently a Professor with the School of Information and Communication Engineering, University of Electronic Science and Technology of China. He has authored or co-authored numerous technical articles in well-known international journals and conferences. His current research interests include image segmentation and object detection. 
\end{IEEEbiography}

\begin{IEEEbiography}
[{\includegraphics[width=1in,height=1.25in,clip,keepaspectratio]{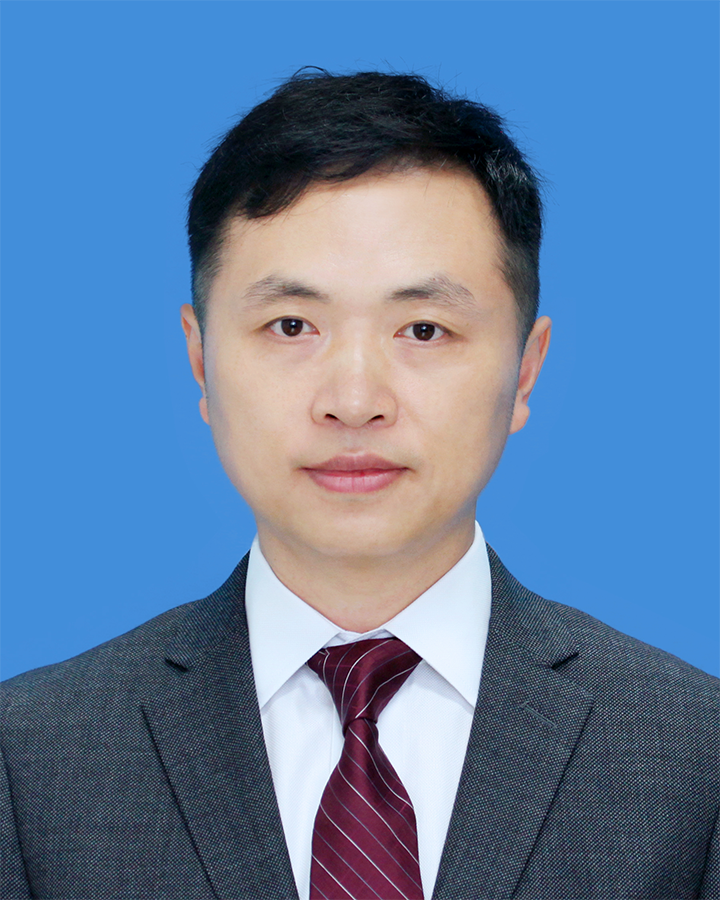}}]{Linfeng Xu} (Member, IEEE) received the Ph.D. degree in Signal and Information Processing from the School of Electronic Engineering, University of Electronic Science and Technology of China (UESTC), Chengdu, China, in 2014. From December 2014 to December 2015, he was with the Ubiquitous Multimedia Laboratory, the State University of New York at Buffalo, USA, as a visiting scholar. He is currently an Associate Professor with the School of Information and Communication Engineering, UESTC. His research interests include machine learning, computer vision, visual signal processing, artificial intelligence theory and applications. He served as a Local Arrangement Chair for ISPACS 2010 and VCIP 2016.
\end{IEEEbiography}

\begin{IEEEbiography}[{\includegraphics[width=1in,height=1.25in,clip,keepaspectratio]{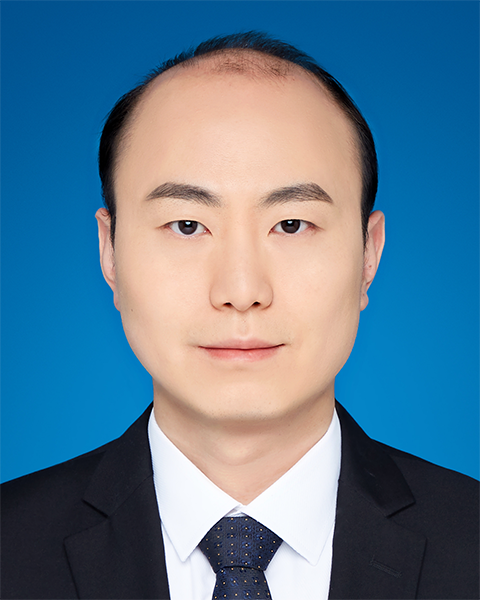}}]{Qingbo Wu} (Member, IEEE) received the Ph.D. degree in signal and information processing from the University of Electronic Science and Technology of China in 2015. From February 2014 to May 2014, he was a Research Assistant with the Image and Video Processing (IVP) Laboratory, Chinese University of Hong Kong. From October 2014 to October 2015, he served as a Visiting Scholar with the Image and Vision Computing (IVC) Laboratory, University of Waterloo. He is currently a Professor with the School of Information and Communication Engineering, University of Electronic Science and Technology of China. His research interests include image/video coding, quality evaluation, perceptual modeling and processing. He has served as Area Chair for ACM MM 2024-2025, VCIP 2016, Session Chair for ACM MM 2021, ICMCT 2022, TPC/PC member of AAAI 2021-2023, APSIPA ASC 2020-2021, CICAI 2021-2023. He was also a Guest Editor of Remote Sensing and Frontiers in Neuroscience. 
\end{IEEEbiography}

\begin{IEEEbiography}[{\includegraphics[width=1in,height=1.25in,clip,keepaspectratio]{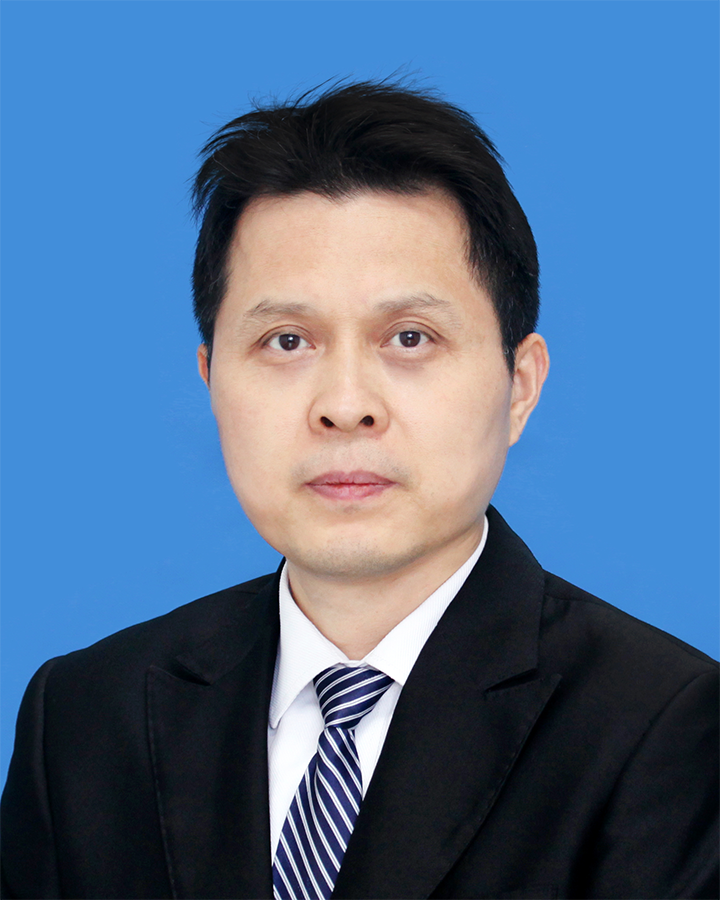}}]{Hongliang Li} (Senior Member, IEEE) received his Ph.D. degree in Electronics and Information Engineering from Xi’an Jiaotong University, China, in 2005. From 2005 to 2006, he joined the visual signal processing and communication laboratory (VSPC) of the Chinese University of Hong Kong (CUHK) as a Research Associate. From 2006 to 2008, he was a Postdoctoral Fellow at the same laboratory in CUHK. He is currently a Professor in the School of Information and Communication Engineering, University of Electronic Science and Technology of China. His research interests include image segmentation, object detection, image and video coding, visual attention, and multimedia processing. 

Dr. Li has authored or co-authored numerous technical articles in well-known international journals and conferences. He is a co-editor of a Springer book titled “Video segmentation and its applications”. Dr. Li is involved in many professional activities. He received the 2019 and 2020 Best Associate Editor Awards for IEEE Transactions on Circuits and Systems for Video Technology (TCSVT), and the 2021 Best Editor Award for Journal on Visual Communication and Image Representation. He served as a Technical Program Chair for VCIP 2016 and PCM 2017, General Chairs for ISPACS 2017 and ISPACS 2010, a Publicity Chair for IEEE VCIP 2013, a Local Chair for the IEEE ICME 2014, Area Chairs for VCIP 2022 and 2021, and a Reviewer committee member for IEEE ISCAS from 2018 to 2022. He served as an Associate Editor of IEEE Transactions on Circuits and Systems for Video Technology (2018-2021). He is now an Associate Editor of Journal on Visual Communication and Image Representation, IEEE Open Journal of Circuits and Systems, and an Area Editor of Signal Processing: Image Communication (Elsevier Science). He is selected as the IEEE Circuits and Systems Society Distinguished Lecturer for 2022-2023.
\end{IEEEbiography}

\end{document}